\documentclass[a4paper, 11pt]{book}
\newcommand{\Website}{http://invibe.net/LaurentPerrinet}%
\newcommand{\Email}{Laurent.Perrinet@univ-amu.fr}%
\newcommand{\Title}{Sparse models for Computer Vision}%
\newcommand{\Acknowledgments}{%
The author was supported by EC IP project FP7-269921, ``BrainScaleS''. 
Correspondence and requests for materials should be addressed to the author\footnote{\Email }.
Code and supplementary material available at \url{\Website/Publications/Perrinet15bicv}. %
}%

\usepackage{siunitx}%
\newcommand{\ms}{\si{\milli\second}}%
\newcommand{\meter}{\si{\meter}}%
\usepackage{graphicx}
\graphicspath{{./figures/}}%
\DeclareGraphicsExtensions{.pdf}
\usepackage{geometry}
\usepackage[round]{natbib}
  {unsrtnat}%
\usepackage[breaklinks]{hyperref}
\author{Perrinet}\title{\Title}
\usepackage{fancyhdr}
\usepackage{tikz}%
\usepackage{tkz-euclide} \usetkzobj{all} % loading all objects
\newcommand{\image}{\mathbf{I}} % the image
\newcommand{\dico}{\Phi} % the dictionary

\newcommand{\coef}{\mathbf{a}} % image's hidden param
\newcommand\la{\leftarrow} %
\newcommand{\sparsenet}{{\sc SparseNet}}%
\newcommand{\eqdef}{\stackrel{\rm def}{=}}%
\newcommand{\seeFig}[1]{Figure~\ref{fig:#1}}%
\newcommand{\seeEq}[1]{Eq.~\ref{eq:#1}}%

\begin{document}
%\include{foreword}
%\tableofcontents
%\include{preface}
\frontmatter
\mainmatter
\addtocounter{chapter}{13} % set them to some other numbers than 0
%\part{}
%\chapterauthor{\AuthorA}
\chapter{\Title}
%=========================================%
%______________%
%\section*{Abstract}
%\Abstract
%__

\paragraph{BibTex entry}~~\\

This chapter appeared as~\citep{Perrinet15sparse}:
\begin{verbatim}

@inbook{Perrinet15sparse,
	Author = {Perrinet, Laurent U.},
	Chapter = {Sparse models for Computer Vision},
	Citeulike-Article-Id = {13514904},
	Editor = {Crist{\'{o}}bal, Gabriel and Keil, Matthias S. and Perrinet, Laurent U.},
	Keywords = {bicv-sparse, sanz12jnp, vacher14},
    month = nov,
	chapter = {14},
	Priority = {0},
	isbn = {9783527680863},
	title = {Sparse Models for Computer Vision},
	DOI = {10.1002/9783527680863.ch14},
    url={http://onlinelibrary.wiley.com/doi/10.1002/9783527680863.ch14/summary},
    publisher = {Wiley-VCH Verlag GmbH {\&} Co. KGaA},
	booktitle = {Biologically inspired computer vision},
	Year = {2015}
	}
\end{verbatim}
\tableofcontents

\section{Motivation}
%----------------------------------------------------%
\subsection{Efficiency and sparseness in biological representations of natural images}
%~~~~~~~~~~~~~~~~~~~~~~~~~~~~~~%
%%------------------------------%
%%: See Figure~\ref{fig:inspiration}
%\begin{figure}%[ht!]%[p!]
%\centering{
%\begin{tikzpicture}%
%\draw (0,0) rectangle (.618\textwidth, .618\textwidth);
%\draw (.618\textwidth, .236\textwidth) rectangle (.382\textwidth, .382\textwidth);
%\draw (.618\textwidth, 0) rectangle (.382\textwidth, .236\textwidth);
%\end{tikzpicture}
%%\includegraphics[width=1.\columnwidth]{inspiration}%
%}
%\caption{ {\bf Sparseness in natural image and in neural activity} (A) TODO : Include an image that is interpretable using sparse principle (occam razor) as a gestalt rule (B) TODO : any image showing sparse activity in V1? Baudot13 (C) sparseness for a single patch
%\label{fig:inspiration}}%
%\end{figure}%
%%------------------------------%
% \showthe\columnwidth

% bio: fundamental question = why sparseness in V1 ? 	 
The central nervous system is a dynamical, adaptive organ
which constantly evolves to provide optimal decisions\footnote{Decisions are defined in their broader sense of elementary choices operated in the system at associative or motor levels~\citep{Friston12}.} for interacting with the environment.
The early visual pathways provides with a powerful system for probing and modeling these mechanisms.
For instance, the primary visual cortex of primates (V1) is absolutely central for most visual tasks.
There, it is observed that some neurons from the input layer of V1 present a selectivity for localized,
edge-like features ---as represented by their ``receptive fields''~\citep{Hubel68}.
Crucially, there is experimental evidence for \emph{sparse} firing in the neocortex~\citep{Barth12,Willmore11}
and in particular in V1.
A representation is sparse when each input signal is associated
with a relatively small sub-set of simultaneously activated neurons within a whole population.
For instance, orientation selectivity of simple cells is sharper
than the selectivity that would be predicted by linear filtering.
Such a procedure produces a rough ``sketch'' of the image
on the surface of V1 that is believed to serve as a ``blackboard''
for higher-level cortical areas~\citep{Marr83}.
However, it is still largely unknown how neural computations act in V1 to represent the image.
More specifically, what is the role of sparseness ---as a generic neural signature---
in the global function of neural computations?

% efficient: information theory, redundancy / lateral interactions / natural
A popular view is that such a population of neurons operates
such that relevant sensory information from the retino-thalamic pathway
is transformed (or ``coded'') efficiently.
Such efficient representation will allow decisions
to be taken optimally in higher-level areas.
In this framework, optimality is defined in terms of information theory~\citep{Attneave54,Atick92,Wolfe10}.
For instance, the representation produced by the neural activity in V1 is sparse:
It is believed that this reduces redundancies
and allows to better segregate edges in the image~\citep{Field94,Froudarakis14}.
% Population code in mouse V1 facilitates readout of natural scenes through increased sparseness  Froudarakis14
This optimization is operated given biological constraints, 
such as the limited bandwidth of information transfer to higher processing stages 
or the limited amount of metabolic resources (energy or wiring length).
More generally it allows to increase the storage capacity of associative memories 
before memory patterns start to interfere with each other~\citep{Palm13}.
Moreover, it is now widely accepted that this redundancy reduction
is achieved in a neural population through lateral interactions.
Indeed, a link between anatomical data and a functional connectivity
between neighboring representations of edges has been found~\citep{Bosking97},
though their conclusions were more recently refined to show that this process may be more complex~\citep{Hunt11}.
By linking neighboring neurons representing similar features, one allows thus a more efficient representation in V1.
As computer vision systems are subject to similar constraints,
applying such a paradigm therefore seems a promising approach towards more biomimetic algorithms.

% meaningful => sparse / more generally predictive coding - link with Gestalt Wertheimer23 => / CV
It is believed that such a property reflects the efficient match of the representation
with the statistics of natural scenes, that is, to behaviorally relevant sensory inputs.
Indeed, sparse representations are prominently observed for cortical responses to natural stimuli~\citep{Field87,Vinje00,Deweese03,Baudot13}.
As the function of neural systems mostly emerges from unsupervised learning, 
it follows that these are adapted to the input which are behaviorally the most common and important.
More generally, by being adapted to natural scenes, this shows that
sparseness is a neural signature of an underlying optimization process.
In fact, one goal of neural computation in low-level sensory areas such as V1
is to provide  relevant predictions~\citep{Rao99,Spratling11}.
This is crucial for living beings as they are often confronted with
noise (internal to the brain or external, such as in low light conditions),
ambiguities (such as inferring a three dimensional slant from a bi-dimensional retinal image).
Also, the system has to compensate for inevitable delays, such as the delay from light stimulation to activation in V1 which is estimated to be of 50~\ms\ in humans.
For instance, a tennis ball moving at 20~\si{\meter\per\second} at one meter in the frontal plane
elicits an input activation in V1 corresponding to  around 45\si{\degree} of visual angle
behind its physical position~\citep{PerrinetAdamsFriston14}. 
Thus, to be able to translate such knowledge to the computer vision community,
it is crucial to better understand \emph{why} the neural processes that produce sparse coding are efficient.
\subsection{Sparseness induces neural organization}
%~~~~~~~~~~~~~~~~~~~~~~~~~~~~~~~~~~~~~~~~~~~~~~~~~%
% breakthrough = learning
A breakthrough in the modeling of the representation in V1 was
the discovery that sparseness is sufficient to induce
the emergence of receptive fields similar to V1 simple cells~\citep{Olshausen96}.
This reflects the fact that, at the learning time scale,
coding is optimized relative to the statistics of natural scenes
such that independent components of the input are represented~\citep{Olshausen97,Bell97}.
The emergence of edge-like simple cell receptive fields in the input layer of area V1 of primates
may thus be considered as a coupled coding and learning optimization problem:
At the coding time scale, the sparseness of the representation is optimized for any given input
while at the learning time scale, synaptic weights are tuned to achieve on average
an optimal representation efficiency over natural scenes.
%%%
This theory has allowed to connect the different fields by providing a link between
information theory models, neuromimetic models and physiological observations. %

% existing models: sparse hebbian learning
In practice, most sparse unsupervised learning models aim at optimizing
a cost defined on prior assumptions on the sparseness of the representation.
These sparse learning algorithms have been applied both for images~\citep{Fyfe95,Olshausen96,Zibulevsky01,Perrinet03ieee,Rehn07,Doi07,Perrinet10shl}
and sounds~\citep{Lewicki00,Smith06}.
Sparse coding may also be relevant to the amount of energy the brain needs to use to sustain its function.
The total neural activity generated in a brain area is
inversely related to the sparseness of the code, therefore the total energy consumption decreases with increasing sparseness.
As a matter of fact the probability distribution functions of neural activity observed experimentally can be approximated by so-called exponential distributions, %  (in bits/spike)
which have the property of maximizing information transmission for a given mean level of activity~\citep{Baddeley97}.
To solve such constraints, some models thus directly compute a sparseness cost based on the representation's distribution.
For instance, the kurtosis corresponds to the 4th statistical moment (the first three moments being in order the mean, variance and skewness)
and  measures how the statistics deviates from a Gaussian:
A positive kurtosis measures if this distribution has an ``heavier tail'' than a Gaussian for a similar variance --- and thus corresponds to a sparser distribution. Based  on such observations, other similar statistical measures of sparseness have been derived in the neuroscience literature~\citep{Vinje00}. % and corresponds to a variant of the above-mentioned sparseness costs.

A more general approach is to derive a representation cost.
For instance, learning is accomplished in the \sparsenet\ algorithmic framework~\citep{Olshausen97}
on image patches taken from natural images as a sequence of coding and learning steps.
First, sparse coding is achieved using a gradient descent over a convex cost. 
We will see later in this chapter how this cost is derived from a prior  on the probability distribution function of the coefficients and how it favors the sparseness of the representation.
At this step, the coding is performed using the current state of the ``dictionary'' of receptive fields.
Then, knowing this sparse solution,
learning is defined as slowly changing the dictionary using Hebbian learning~\citep{Hebb49}.
As we will see later, the parameterization of the prior
has a major impact on the results of the sparse coding and
thus on the emergence of edge-like receptive fields and requires proper tuning.
Yet, this class of models provides a simple solution to the problem of sparse representation in V1.

% neural signatures
However, these models are quite abstract and assume that neural computations may estimate some rather complex measures such as gradients - a problem that may also be faced by neuromorphic systems.
Efficient, realistic implementations have been proposed which show that imposing sparseness
may indeed guide neural organization in neural network models, see for instance~\citep{Zylberberg11,Hunt13}.
Additionally, it has also been shown that in a neuromorphic model,
an efficient coding hypothesis links sparsity and selectivity of neural responses~\citep{Blattler11}.
More generally, such neural signatures are reminiscent of the shaping of neural activity to account for contextual influences.
For instance, it is observed that ---depending on the context outside the receptive field of a neuron in area V1--- the tuning curve may demonstrate a modulation of its orientation selectivity.
This was accounted for instance as a way to optimize the coding efficiency of a population of neighboring neurons~\citep{Series04}.
As such, sparseness is a relevant neural signature for a large class of neural computations implementing efficient coding.
\subsection{Outline: Sparse models for Computer Vision}
%~~~~~~~~~~~~~~~~~~~~~~~~~~~~~~~~~~~~~~~~~~~~~~~~~~~~~%
% sparse models: other approaches
As a consequence, sparse models provide a fruitful approach for computer vision.
It should be noted that other popular approaches for taking advantage of sparse representations exist.
The most popular is compressed sensing~\citep{Ganguli12},
for which it has been proven that ---assuming sparseness in the input,
it is possible to reconstruct the input from a sparse choice
of linear coefficients computed from randomly drawn basis functions.
Note also that some studies also focus on temporal sparseness.
Indeed, by computing for a given neuron
the relative numbers of active events relative to a given time window,
one computes the so-called lifetime sparseness (see for instance~\citep{Willmore11}).
We will see below that this measure may be related to population sparseness.
For a review of sparse modeling approaches, we refer to~\citep{Elad10}.
Herein, we will focus on the particular sub-set of such models based on their biological relevance.

% intelligent machines? Neuroscience is beginning to inspire a new generation of seeing machines.
Indeed, we will rather focus on biomimetic sparse models as tools to shape future computer vision algorithms~\citep{Benoit10,Serre10}.
In particular, we will not review models which mimic neural activity,
but rather on algorithms which mimic their efficiency,
bearing in mind  the constraints that are linked to neural systems (no central clock, internal noise, parallel processing, metabolic cost, wiring length).
For that purpose, we will complement some previous studies~\citep{Perrinet03ieee,Fischer07,Perrinet08spie,Perrinet10shl} (for a review see~\citep{Perrinet07})
by putting these results in light of most recent theoretical and physiological findings.

% outline for a CV audience
This chapter is organized as follows.
First, in Section~\ref{sec:sparsenet} we will outline how we may implement the unsupervised learning algorithm at a local scale for image patches.
Then we will extend in Section~\ref{sec:sparselet} such an approach to full scale natural images by defining the \emph{SparseLets} framework.
Such formalism will then be extended in Section~\ref{sec:sparseedges} to include context modulation, for instance from higher-order areas.
These different algorithms (from the local scale of image patches to more global scales)
will each be accompanied by a supporting implementation (with the source code)
for which we will show example usage and results.
We will in particular highlight novel results and then draw some conclusions on the perspective of sparse models for computer vision.
More specifically, we will propose that bio-inspired approaches
may be applied to computer vision using predictive coding schemes,
sparse models being one simple and efficient instance of such schemes.
\section{What is sparseness? Application to image patches}
%--------------------------------------------------------%
\label{sec:sparsenet}
\subsection{Definitions of sparseness}
%~~~~~~~~~~~~~~~~~~~~~~~~~~~~~~~~~~~~%
%from a measure of efficiency...
In low-level sensory areas, the goal of neural computations is to generate efficient intermediate \emph{representations} as we have seen that this allows more efficient decision making.
Classically, a representation is defined as the inversion of an internal generative model of the sensory world,
that is, by inferring the sources that generated the input signal.
Formally, as in~\citep{Olshausen97}, we define a Generative Linear Model (GLM)
for describing natural, static, grayscale image patches $\image$ (represented by column vectors of dimension $L$ pixels),
by setting a ``dictionary'' of $M$ images (also called ``atoms'' or ``filters'') as the $L \times M$ matrix $\dico=\{ \dico_i\}_{1\leq i \leq M}$.
Knowing the associated ``sources'' as a vector of coefficients $\coef=\{ a_i \}_{1\leq i \leq M}$, the image is defined using matrix notation as a sum of weighted atoms:%
\begin{equation}%
\image = \dico\coef + \mathbf{n}%
\label{eq:lgm}%
\end{equation} %
where $\mathbf{n}$ is a Gaussian additive noise image. This noise, as in~\citep{Olshausen97}, is scaled to a variance of $\sigma_n^2$ to achieve decorrelation by applying Principal Component Analysis to the raw input images, without loss of generality since this preprocessing is invertible. Generally, the dictionary $\dico$ may be much larger than the dimension of the input space (that is, if $M \gg L$) and it is then said to be \emph{over-complete}. However, given an over-complete dictionary, the inversion of the GLM leads to a combinatorial search and typically, there may exist many coding solutions $\coef$ to \seeEq{lgm} for one given input $\image$. The goal of efficient coding is to find, given the dictionary $\dico$ and for any observed signal $\image$, the ``best'' representation vector, that is, as close as possible to the sources that generated the signal. Assuming that for simplicity, each individual coefficient is represented in the neural activity of a single neuron, this would justify the fact that this activity is sparse. It is therefore necessary to define an efficiency criterion in order to choose between these different solutions. % (see Figure~\ref{fig:inspiration}-B)

% ... to sparseness
Using the GLM, we will infer the ``best'' coding vector as the most probable. In particular, from the physics of the synthesis of natural images, we know \emph{a priori} that image representations are sparse: They are most likely generated by a small number of features relatively to the dimension $M$ of the representation space. % (see Figure~\ref{fig:inspiration}-C).
Similarly to \citet{Lewicki00}, this can be formalized in the probabilistic framework defined by the GLM (see~\seeEq{lgm}), by assuming that knowing the prior distribution of the coefficients $a_i$ for natural images, the representation cost of $\coef$ for one given natural image is:%
\begin{eqnarray}%
\mathcal{C}( \coef | \image , \dico) 	&\eqdef& -\log P( \coef | \image , \dico) 
=  \log P( \image )  -\log P( \image | \coef , \dico) -  \log P(\coef | \dico) \nonumber \\
&=& \log P( \image ) + \frac{1}{2\sigma_n^2} \| \image - \dico \coef \|^2 - \sum_i \log P(a_i | \dico)%
\label{eq:efficiency_cost}%
\end{eqnarray}%
where $P( \image )$ is the partition function which is independent of the coding (and that we thus ignore in the following) and $ \| \cdot \|$ is the L$_2$-norm in image space. This efficiency cost is measured in bits if the logarithm is of base 2, as we will assume without loss of generality thereafter. For any representation $\coef$, the cost value corresponds to the description length~\citep{Rissanen78}: On the right hand side of~\seeEq{efficiency_cost}, the second term corresponds to the information from the image which is not coded by the representation (reconstruction cost) and thus to the information that can be at best encoded using entropic coding pixel by pixel (that is, the negative log-likelihood $-\log P( \image | \coef , \dico)$ in Bayesian terminology, see chapter \verb+009_series+ for Bayesian models applied to computer vision). The third term $S( \coef | \dico) = - \sum_i \log P(a_i | \dico)$ is the representation or sparseness cost: It quantifies representation efficiency as the coding length of each coefficient of $\coef$ which would be achieved by entropic coding knowing the prior and assuming that they are independent.
The rightmost penalty term (see Equation~\ref{eq:efficiency_cost}) gives thus a definition of sparseness $S( \coef | \dico)$ as the sum of the log prior of coefficients.

In practice, the sparseness of coefficients for natural images is often defined by an \emph{ad hoc} parameterization of the shape of the prior. For instance, the parameterization in~\citet{Olshausen97} yields the coding cost:%
\begin{equation}
\mathcal{C}_1( \coef | \image , \dico) =\frac{1}{2\sigma_n^2} \| \image - \dico \coef \|^2 + \beta \sum_i \log ( 1 + \frac{a_i^2}{\sigma^2} )%
\label{eq:sparse_cost}%
\end{equation}
where $\beta$ corresponds to the steepness of the prior and $\sigma$ to its scaling (see Figure 13.2 from~\citep{Olshausen02}). This choice is often favored because it results in a convex cost for which known numerical optimization methods such as conjugate gradient may be used. In particular, these terms may be put in parallel to regularization terms that are used in computer vision. For instance, a L2-norm penalty term corresponds to Tikhonov regularization~\citep{Tikhonov77} or a L1-norm term corresponds to the Lasso method. 
See chapter \verb+003_holly_gerhard+ for a review of possible parametrization of this norm, 
for instance by using nested $L_p$ norms.
Classical implementation of sparse coding rely therefore on a parametric measure of sparseness.

% $P(s_i)\approx cste$ and $P(s_i=0)\approx 1$ leads to L0
Let's now derive another measure of sparseness.
Indeed, a non-parametric form of sparseness cost may be defined by considering that neurons representing the vector $\coef$ are either active or inactive.
In fact, the spiking nature of neural information demonstrates that the transition from an inactive to an active state is far more significant at the coding time scale than smooth changes of the firing rate. This is for instance perfectly illustrated by the binary nature of the neural code in the auditory cortex of rats~\citep{Deweese03}. Binary codes also emerge as optimal neural codes for rapid signal transmission~\citep{Bethge03,Nikitin09}. This is also relevant for neuromorphic systems which transmit discrete events (such as a network packet). With a binary event-based code, the cost is only incremented when a new neuron gets active, regardless to the analog value. Stating that an active neuron carries a bounded amount of information of $\lambda$ bits, an upper bound for the representation cost of neural activity on the receiver end is proportional to the count of active neurons, that is, to the $\ell_0$ pseudo-norm $\| \coef \|_0$:%
\begin{equation}%
\mathcal{C}_0( \coef | \image , \dico) = \frac{1}{2\sigma_n^2} \| \image - \dico \coef \|^2 + \lambda \| \coef \|_0%
\label{eq:L0_cost}%
\end{equation}%
This cost is similar with information criteria such as the Akaike Information Criteria~\citep{Akaike74} or distortion rate~\citep[p.~571]{Mallat98}. This simple non-parametric cost has the advantage of being dynamic: The number of active cells for one given signal grows in time with the number of spikes reaching the target population. But \seeEq{L0_cost} defines a harder cost to optimize (in comparison to Equation~\ref{eq:sparse_cost} for instance) since the hard $\ell_0$ pseudo-norm sparseness leads to a non-convex optimization problem which is \emph{NP-complete} with respect to the dimension $M$ of the dictionary~\citep[p.~418]{Mallat98}.%

\subsection{Learning to be sparse: the SparseNet algorithm}
%~~~~~~~~~~~~~~~~~~~~~~~~~~~~~~~~~~~~~~~~~~~~~~~~~~~~~~~~~~~~~~~%
We have seen above that we may define different models for measuring sparseness depending on our prior assumption on the distribution of coefficients.
Note first that, assuming that the statistics are stationary (more generally ergodic), then these measures of sparseness across a population should necessary imply a lifetime sparseness for any neuron.
Such a property is essential to extend results from electro-physiology.
Indeed, it is easier to record a restricted number of cells
than a full population (see for instance~\citep{Willmore11}).
However, the main property in terms of efficiency  is that the representation should be sparse at any given time,
that is, in our setting, at the presentation of each novel image.

Now that we have defined sparseness, how could we use it to induce neural organization?
Indeed, given a sparse coding strategy that optimizes any representation efficiency cost as defined above, we may derive an unsupervised learning model by optimizing the dictionary $\dico$ over natural scenes. On the one hand, the flexibility in the definition of the sparseness cost leads to a wide variety of proposed \emph{sparse coding} solutions (for a review, see~\citep{Pece02}) such as numerical optimization~\citep{Olshausen97}, non-negative matrix factorization~\citep{Lee99,Ranzato07} or Matching Pursuit~\citep{Perrinet03ieee,Smith06,Rehn07,Perrinet10shl}. They are all derived from correlation-based inhibition since this is necessary to remove redundancies from the linear representation. This is consistent with the observation that lateral interactions are necessary for the formation of elongated receptive fields~\citep{Bolz89,Wolfe10}.

On the other hand, these methods share the same GLM model (see~\seeEq{lgm}) and once the sparse coding algorithm is chosen, the learning scheme is similar.
As a consequence, after every coding sweep, we increased the efficiency of the dictionary $\dico$  with respect to \seeEq{efficiency_cost}. This is achieved using  the online gradient descent approach given the current sparse solution,  $\forall i$:
\begin{equation}%
\dico_{i} \la \dico_{i} + \eta \cdot a_{i} \cdot (\image - \dico\coef)%
\label{eq:learn}%
\end{equation}%
where $\eta$ is the learning rate. %
Similarly to Eq.~17 in~\citep{Olshausen97} or to Eq.~2 in~\citep{Smith06}, the relation is a linear ``Hebbian'' rule~\citep{Hebb49} since it enhances the weight of neurons proportionally to the correlation between pre- and post-synaptic neurons. Note that there is no learning for non-activated coefficients. The novelty of this formulation compared to other linear Hebbian learning rule such as~\citep{Oja82} is to take advantage of the sparse representation, hence the name Sparse Hebbian Learning (SHL).%

The class of SHL algorithms are unstable without homeostasis, that is, without a process that maintains the system in a certain equilibrium. In fact, starting with a random dictionary, the first filters to learn are more likely to correspond to salient features~\citep{Perrinet03ieee} and are therefore more likely to be selected again in subsequent learning steps. In {\sc SparseNet}, the homeostatic gain control is implemented by adaptively tuning the norm of the filters. This method equalizes the variance of coefficients across neurons using a geometric stochastic learning rule. The underlying heuristic is that this introduces a bias in the choice of the active coefficients. In fact, if a neuron is not selected often, the geometric homeostasis will decrease the norm of the corresponding filter, and therefore ---from~\seeEq{lgm} and the conjugate gradient optimization--- this will increase the value of the associated scalar. Finally, since the prior functions defined in~\seeEq{sparse_cost} are identical for all neurons, this will increase the relative probability that the neuron is selected with a higher relative value. The parameters of this homeostatic rule have a great importance for the convergence of the global algorithm. In~\citep{Perrinet10shl}, we have derived a more general homeostasis mechanism derived from the optimization of the representation efficiency through histogram equalization which we will describe later (see Section~\ref{sec:laughlin}).
%------------------------------%
%: See Figure~\ref{fig:sparsenet}
\begin{figure}%[ht!]%[p!]
\centering{%
\begin{tikzpicture}%[scale=1, font=\sffamily]%,every node/.style={minimum size=1cm},on grid]
\draw [anchor=north west] (.33\textwidth, 0) node {\includegraphics[width=.33\textwidth]{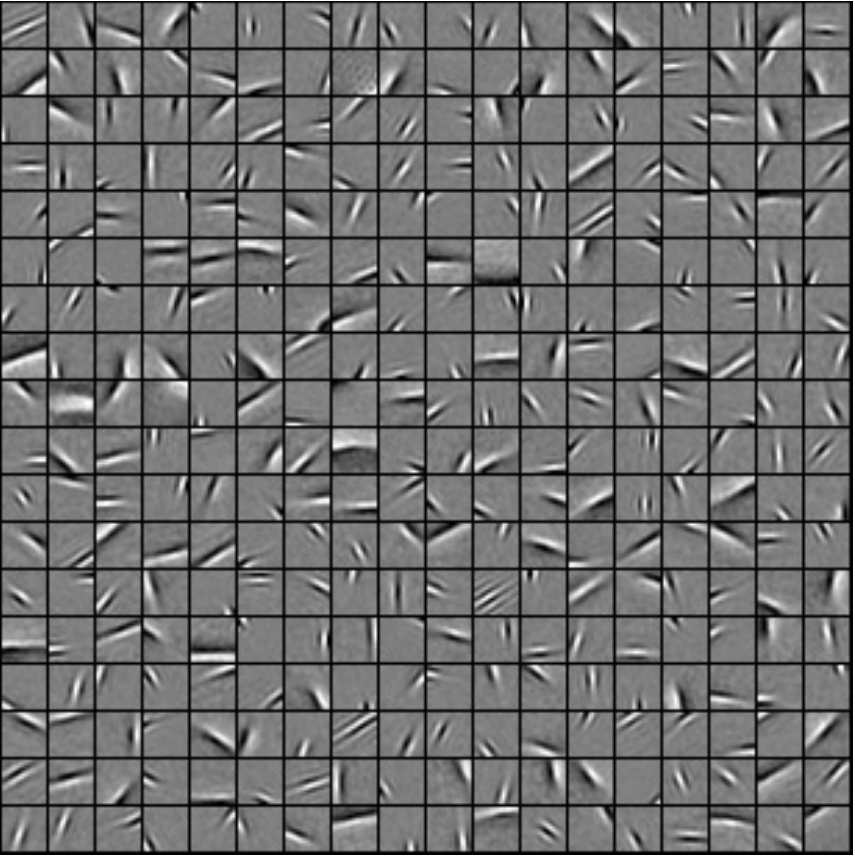}};
\draw [anchor=north west] (.66\textwidth, 0) node {\includegraphics[width=.33\textwidth]{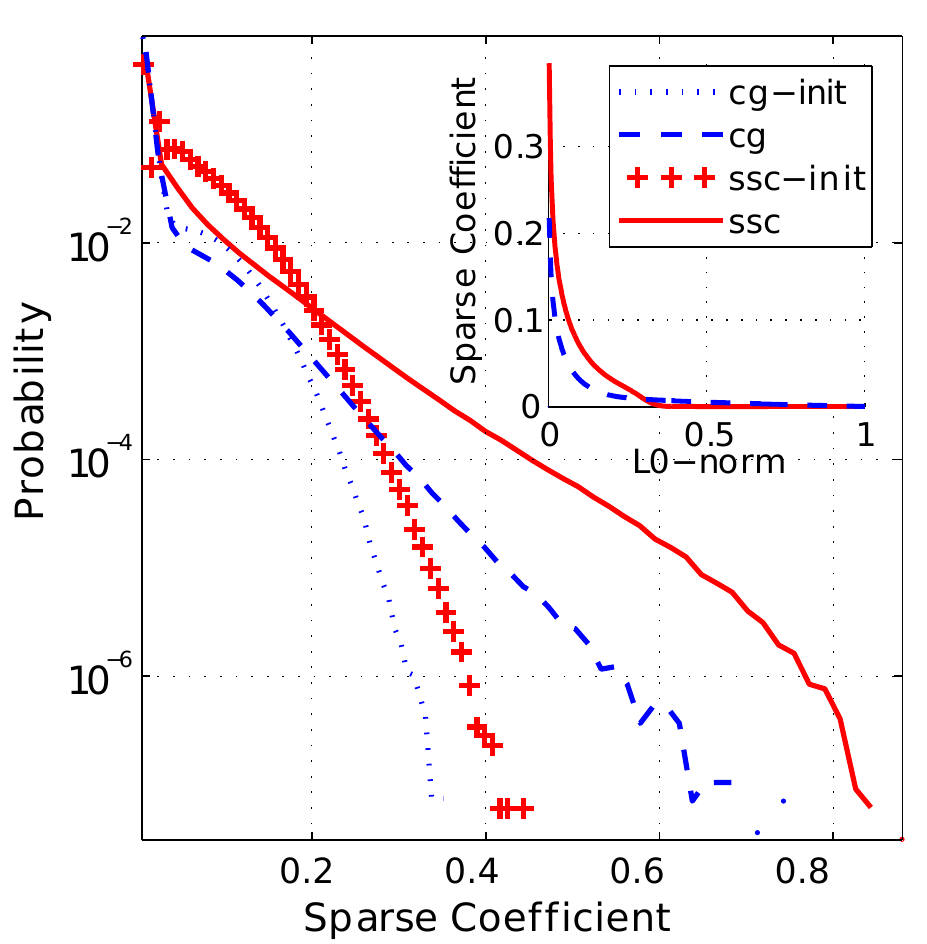}};
\draw [anchor=north west] (.99\textwidth, 0) node {\includegraphics[width=.3275\textwidth]{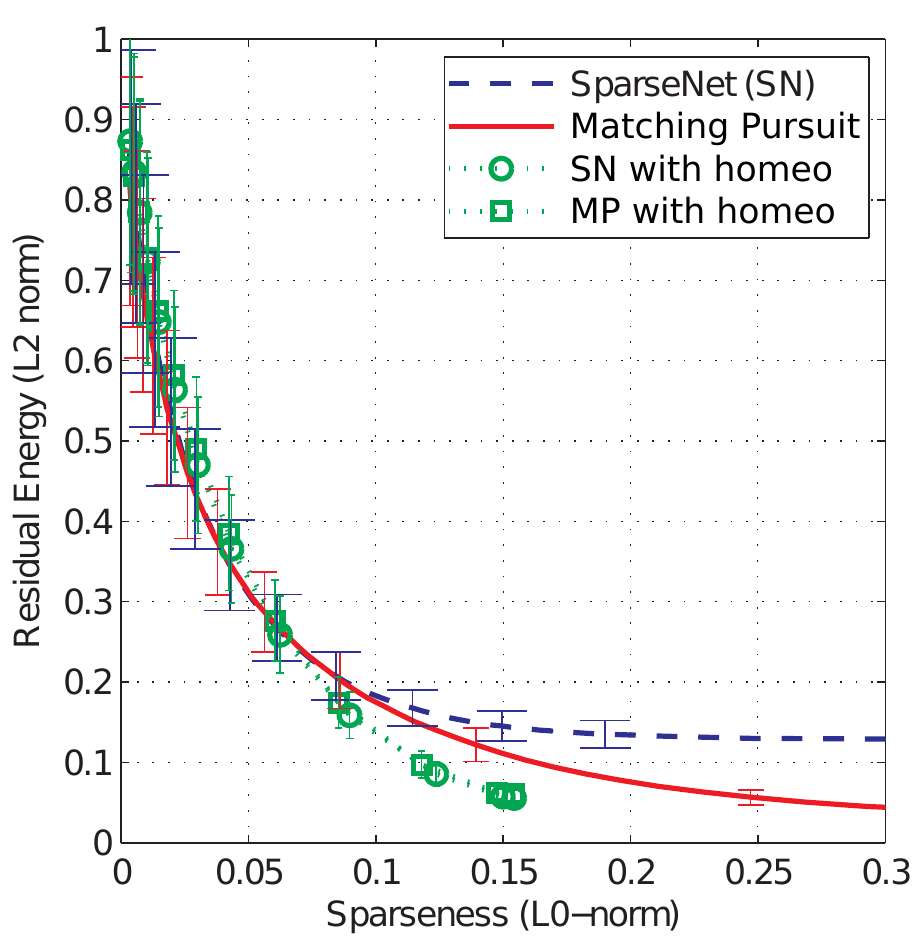}};
\draw [anchor=north west] (.33\textwidth, .03\textwidth) node {$\mathsf{(A)}$};
\draw [anchor=north west] (.66\textwidth, .03\textwidth) node {$\mathsf{(B)}$};
\draw [anchor=north west] (.99\textwidth, .03\textwidth) node {$\mathsf{(C)}$};
\end{tikzpicture}%
}
\caption{ {\bf Learning a sparse code using Sparse Hebbian Learning.}  \textsf{(A)}~We show the results  at convergence (20000 learning steps) of a sparse model with unsupervised learning algorithm which progressively optimizes the relative number of active (non-zero) coefficients ($\ell_0$ pseudo-norm)~\citep{Perrinet10shl}. Filters of the same size as the image patches are presented in a matrix (separated by a black border). Note that their position in the matrix is arbitrary as in ICA. These results show that sparseness induces the emergence of edge-like receptive fields similar to those observed in the primary visual area of primates. \textsf{(B)}~We show the probability distribution function of sparse coefficients obtained by our method compared to~\citep{Olshausen96} with first, random dictionaries (respectively 'ssc-init' and 'cg-init') and second, with the dictionaries obtained after convergence of respective learning schemes (respectively 'ssc' and 'cg'). At convergence, sparse coefficients are more sparsely distributed than initially, with more kurtotic probability distribution functions for {\sf 'ssc'} in both cases, as can be seen in the ``longer tails''of the distribution. \textsf{(C)}~We evaluate the coding efficiency of both methods by plotting the average residual error (L$_2$ norm) as a function of the $\ell_0$ pseudo-norm. This provides a measure of the coding efficiency for each dictionary over the set of image patches (error bars represent one standard deviation). Best results are those providing a lower error for a given sparsity (better compression) or a lower sparseness for the same error. %  (similarly to Occam's razor). %
\label{fig:sparsenet}}%
\end{figure}
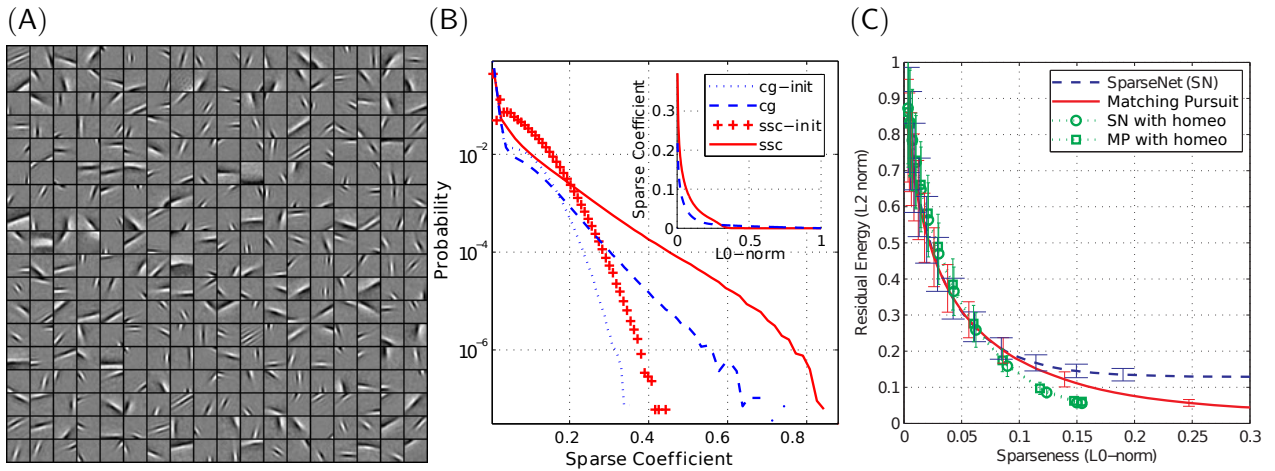%
%------------------------------%
\subsection{Results: efficiency of different learning strategies}
%~~~~~~~~~~~~~~~~~~~~~~~~~~~~~~~~~~~~~~~~~~~~~~~~~~~~~~~~~~~~~~~%
The different SHL algorithms simply differ by the coding step.
This implies that they only differ by first, how sparseness is defined at a functional level and second, how the inverse problem corresponding to the coding step is solved at the algorithmic level.
Most of the schemes cited above use a less strict, parametric definition of sparseness (like the convex L$_1$-norm), but for which a mathematical formulation of the optimization problem exists.
Few studies such as~\citep{Liu14,Peharz12} use the stricter $\ell_0$ pseudo-norm as the coding problem gets more difficult.
A thorough comparison of these different strategies was recently presented in~\citep{Charles12}.
See also~\citep{Aharon06} for properties of the coding solutions to the $\ell_0$ pseudo-norm.
Similarly, in~\citep{Perrinet10shl}, we preferred to retrieve an approximate solution to the coding problem to have a better match with the measure of efficiency~\seeEq{L0_cost}. %  (see Section~\ref{sec:matchingpursuit} for a description of the algorithm)

Such an algorithmic framework is implemented in the SHL-scripts package\footnote{These scripts are available at~\url{https://github.com/bicv/SHL_scripts} and documented at~\url{https://pythonhosted.org/SHL_scripts}.}.
These scripts allow the retrieval of the database of natural images and the replication of the results of~\citep{Perrinet10shl} reported in this section. With a correct tuning of parameters, we observed that different coding schemes show qualitatively a similar emergence of edge-like filters. The specific coding algorithm used to obtain this sparseness appears to be of secondary importance as long as it is adapted to the data and yields sufficiently efficient sparse representation vectors. However, resulting dictionaries vary qualitatively among these schemes and it was unclear which algorithm is the most efficient and what was the individual role of the different mechanisms that constitute SHL schemes. At the learning level, we have shown that the homeostasis mechanism had a great influence on the qualitative distribution of learned filters~\citep{Perrinet10shl}.

Results are shown in Figure~\ref{fig:sparsenet}.
This figure represents the qualitative results of the formation of edge-like filters (receptive fields).
More importantly, it shows the quantitative results as the average decrease of the squared error as a function of the sparseness.
This gives a direct access to the cost as computed in Equation~\ref{eq:L0_cost}.
These results are comparable with the \sparsenet\ algorithm.
Moreover, this solution,
by giving direct access to the atoms (filters) that are chosen,
provides with a more direct tool to manipulate sparse components.
One further advantage consists in the fact that this unsupervised learning model is non-parametric (compare with~\seeEq{sparse_cost})
and thus does not need to be parametrically tuned.
Results show the role of homeostasis on the unsupervised algorithm.
In particular, using the comparison of coding and decoding efficiency with and without this specific homeostasis,
we have proven that cooperative homeostasis optimized overall representation efficiency (see also Section~\ref{sec:laughlin}).

It is at this point important to note that in this algorithm,
we achieve an exponential convergence of the squared error~\citep[p.~422]{Mallat98},
but also that this curve can be directly derived from the coefficients' values.
Indeed, for $N$ coefficients (that is, $\| \coef \|_0 = N$), we have the squared error equal to:
\begin{equation}%
E_N \eqdef \| \image - \dico\coef  \| ^2  / \| \image \| ^2 =  1 - \sum_{1\leq k\leq N} a_{k}^2 / \| \image \| ^2%
\label{eq:error}%
\end{equation}%
As a consequence, the sparser the distributions of coefficients, then quicker is the decrease of the residual energy.
%Note that the observed distribution of coefficients follow a power-law. This was already observed in~\citep{Perrinet03ieee}. This power-law (``scale-free'') distribution is defined by
%\begin{equation}%
%\log p(a) \propto - \gamma \log a_{k}%
%\label{eq:powerlaw}%
%\end{equation}%
%The value of $\gamma$ quantifies therefore the strength of the sparseness in the signal.
% TODO: discuss this? or say it will be discussed below
In the following section, we will describe different variations of this algorithm. To compare their respective efficiency,
we will plot the decrease of the coefficients along with the decrease of the residual's energy. % as a measure of sparseness.
Using such tools, we will now explore if such a property extends to full-scale images and not only to image patches,
an important condition for using sparse models in computer vision algorithms.
% and we will explore this aspect in the following
%
\section{SparseLets: a multi scale, sparse, biologically inspired representation of natural images}
%-------------------------------------------------------------------------------------------------%
\label{sec:sparselet}
\subsection{Motivation: architecture of the primary visual cortex}
%~~~~~~~~~~~~~~~~~~~~~~~~~~~~~~~~~~~~~~~~~~~~~~~~~~~~~~~~~~~~~~~~%
% sparse models for CV / sparse models for image patches / extend it to full images = more sparseness / but complex : let's focus on coding - V1 inspired
Our goal here is to build practical algorithms of sparse coding for computer vision.
We have seen above that it is possible to build an adaptive model of sparse coding
that we applied to $12 \times 12$ image patches.
Invariably, this has shown that the independent components of image patches are edge-like filters,
such as is found in simple cells of V1.
This model has shown that for randomly chosen image patches,
these may be described by a sparse vector of coefficients.
Extending this result to full-field natural images,
we can expect that this sparseness would increase by a degree of order.
In fact, except in a densely cluttered image such as a close-up of a texture,
natural images tend to have wide areas which are void (such as the sky, walls or uniformly filled areas).
However, applying directly the \sparsenet\ algorithm to full-field images is impossible in practice
as its computer simulation would require too much memory to store the over-complete set of filters.
However, it is still possible to define \emph{a priori} these filters and herein,
we will focus on a full-field sparse coding method
whose filters are inspired by the architecture of the primary visual cortex.

\begin{figure}
\centering{%
\begin{tikzpicture}%[scale=1, font=\sffamily]%,every node/.style={minimum size=1cm},on grid]
\node [anchor=north west] (b) at (0, .382\textwidth) {\includegraphics[width=.382\textwidth]{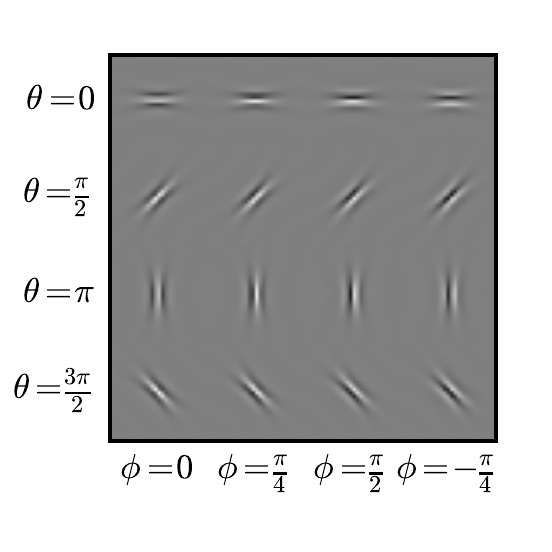}};
\node [anchor=north west] (gp) at (.382\textwidth, .38\textwidth) {\includegraphics[width=.605\textwidth]{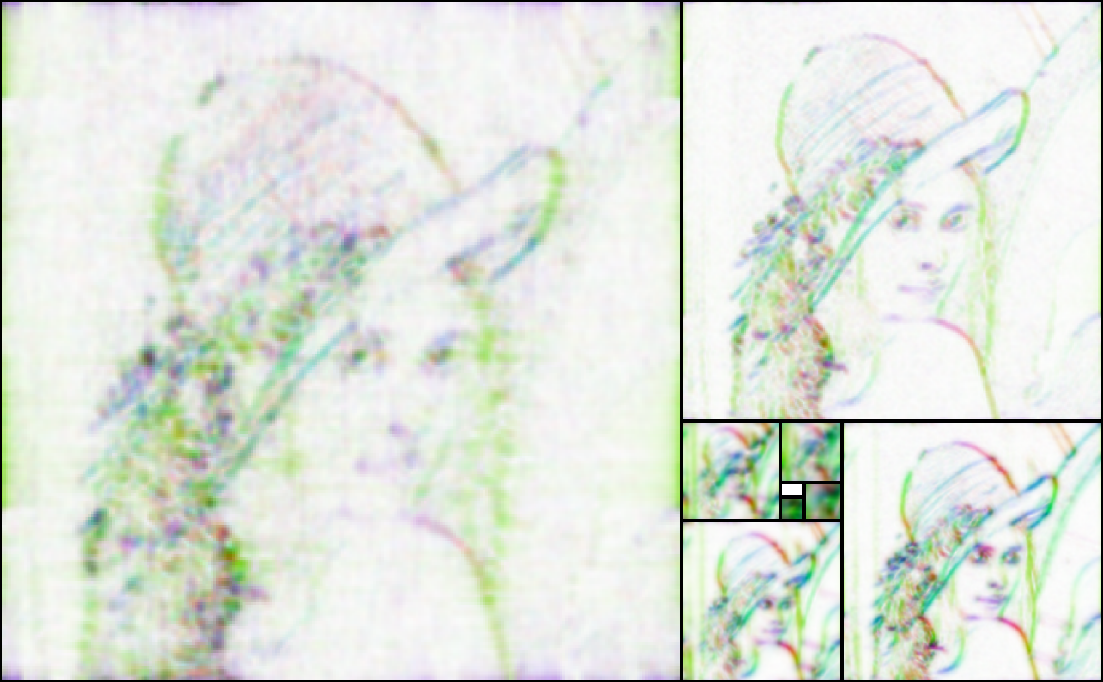}};
\draw [anchor=south west] (0, .382\textwidth) node {$\mathsf{(A)}$};
\draw [anchor=south west] (.382\textwidth, .382\textwidth) node {$\mathsf{(B)}$};
\end{tikzpicture}
}
\caption{ {\bf The log-Gabor pyramid}
\textsf{(A)}~A set of log-Gabor filters showing first in different rows different orientations and second in different columns different phases.
Here we have only shown one scale.
Note the similarity with Gabor filters.
\textsf{(B)}~Using this set of filter, one can define a linear representation that is rotation-, scaling- and translation- invariant.
Here we show a tiling of the different scales according to a Golden Pyramid~\citep{Perrinet08spie}.
The hue gives the orientation while the value gives the absolute value (white denotes a low coefficient).
Note the redundancy of the linear representation, for instance at different scales.
\label{fig:loggabor}}%
\end{figure}

% see http://www.csse.uwa.edu.au/~pk/research/matlabfns/PhaseCongruency/Docs/convexpl.html
The first step of our method involves defining the dictionary of templates (or
filters) for detecting edges.
We use a log-Gabor representation, which is well suited to represent
a wide range of natural images~\citep{Fischer07}.
This representation gives a generic model of edges parameterized by their shape,
orientation, and scale. We set the range of these parameters to match with what has been reported
for simple-cell responses in macaque primary visual cortex (V1).
Indeed log-Gabor filters are similar to standard Gabors and both are well fitted to V1 simple cells~\citep{Daugman80}.
Log-Gabors are known to produce a sparse set of linear coefficients~\citep{Field99}.
Like Gabors, these filters are defined by Gaussians in Fourier space,
but their specificity is that log-Gabors have Gaussians envelopes in log-polar frequency space.
This is consistent with physiological measurements which indicate that V1 cell responses are symmetric on the log frequency scale.
They have multiple advantages over Gaussians:
In particular, they have no DC component,
and more generally, their envelopes more broadly cover the frequency space~\citep{Fischer07cv}.
In this chapter, we set the bandwidth of the Fourier representation of the filters
to $0.4$ and $\pi/8$ respectively in the log-frequency and polar coordinates
to get a family of relatively elongated (and thus selective) filters
(see~\citet{Fischer07cv} and Figure~\ref{fig:loggabor}-A for examples of such edges).
Prior to the analysis of each image, we used the spectral whitening filter
described by~\citet{Olshausen97} to provide
a good balance of the energy of output coefficients~\citep{Perrinet03ieee,Fischer07}.
Such a representation is implemented in the \verb+LogGabor+ package
\footnote{Python scripts are available at \url{https://github.com/bicv/LogGabor}
and documented at \url{https://pythonhosted.org/LogGabor}.}.
%see also hansen and hess 06 for the use of such filters

This transform is linear and can be performed by a simple convolution
repeated for every edge type.
Following~\citet{Fischer07cv}, convolutions were performed in the Fourier (frequency) domain
for computational efficiency.
The Fourier transform allows for a convenient definition of the edge filter characteristics,
and convolution in the spatial domain is equivalent to a simple multiplication in the frequency domain.
By multiplying the envelope of the filter and the Fourier transform of the image,
one may obtain a filtered spectral image that may be converted to
a filtered spatial image using the inverse Fourier transform.
We exploited the fact that by omitting the symmetrical lobe of the envelope of the filter
in the frequency domain,
we obtain quadrature filters.
Indeed, the output of this procedure gives a complex number whose
real part corresponds to the response to the symmetrical part of the edge,
while the imaginary part corresponds to the asymmetrical part of the edge
(see~\citet{Fischer07cv} for more details).
More generally, the modulus of this complex number gives the energy response
to the edge ---as can be compared to the response of complex cells in area V1,
while its argument gives the exact phase of the filter (from symmetric to non-symmetric).
This property further expands the richness of the representation.

Given a filter at a given orientation and scale,
a linear convolution model provides a translation-invariant representation.
Such invariance can be extended to rotations and scalings
by choosing to multiplex these sets of filters at different orientations and spatial scales.
Ideally, the parameters of edges would vary in a continuous fashion,
to a full relative translation, rotation, and scale invariance.
However this is difficult to achieve in practice and some compromise has to be found.
Indeed, though orthogonal representations are popular in computer vision
due to their computational tractability,
it is desirable in our context that we have a relatively high over-completeness
in the representation to achieve this invariance.
For a given set of $256\times 256$ images, we first chose to have $8$ dyadic levels (that is, by doubling the scale at each level)
 with $24$ different orientations.
Orientations are measured as a non-oriented angle in radians, by convention
in the range from $-\frac{\pi}{2}$ to $\frac{\pi}{2}$ (but not including $-\frac{\pi}{2}$) with respect to the $x-$axis.
Finally, each image is transformed into a pyramid of coefficients.
This pyramid consists of approximately $4/3\times256^{2}\approx8.7\times10^4$ pixels
multiplexed on $8$ scales and $24$ orientations, that is,
approximately $16.7\times10^6$ coefficients, an over-completeness factor of about $256$.
This linear transformation is represented by a pyramid of coefficient, see Figure~\ref{fig:loggabor}-B.
%
%------------------------------%
%: See Figure~\ref{fig:loggabor}
%------------------------------%
\subsection{The SparseLets framework}
%~~~~~~~~~~~~~~~~~~~~~~~~~~~~~~~~~~~~~~~~~~%
\label{sec:matchingpursuit}
% over-completeness / combinatorial / bio-inspired = matching pursuit
The resulting dictionary of edge filters is over-complete.
The linear representation would thus give a dense, relatively inefficient representation of the distribution of edges, see Figure~\ref{fig:loggabor}-B.
Therefore, starting from this linear representation,
we searched instead for the most sparse representation.
As we saw above in Section~\ref{sec:sparsenet}, minimizing the $\ell_0$ pseudo-norm (the number of non-zero coefficients) leads
to an expensive combinatorial search with regard to the dimension of the dictionary (it is NP-hard).
As proposed by~\citet{Perrinet03ieee}, we may approximate a
solution to this problem using a greedy approach.
Such an approach is based on the physiology of V1.
Indeed, it has been shown that inhibitory interneurons decorrelate excitatory cells to drive sparse code formation~\citep{Bolz89,King13}.
We use this local architecture to iteratively modify the linear representation~\citep{Fischer07}.

In general, a greedy approach is applied when the optimal combination is
difficult to solve globally, but can be solved progressively,
one element at a time.
Applied to our problem, the greedy approach corresponds to first choosing
the single filter $\Phi_i$ that best fits the image along with a suitable coefficient $a_i$,
such that the single source $a_i\Phi_i$ is a good match to the image.
Examining every filter $\Phi_j$, we find the filter $\Phi_i$
with the maximal correlation coefficient (``Matching'' step), where:
\begin{equation}
i = \mbox{argmax}_j \left( \left\langle \frac{\mathbf{I}}{\| \mathbf{I} \|} , \frac{
\Phi_j}{\| \Phi_j\|} \right\rangle \right),
\label{eq:coco}
\end{equation}
$\langle \cdot,\cdot \rangle$ represents the inner product, and $\| \cdot \|$
represents the $L_2$ (Euclidean) norm.
The index (``address'') $i$ gives the position ($x$ and $y$), scale and orientation of the edge.
We saw above that since filters at a given scale and orientation are generated by a translation,
this operation can be efficiently computed using a convolution,
but we keep this notation for its generality.
The associated coefficient is the scalar projection:
\begin{equation}
a_{i} = \left\langle \mathbf{I} , \frac{ \Phi_{i}}{\| \Phi_{i}\|^2} \right\rangle
\label{eq:proj}
\end{equation}
Second, knowing this choice, the image can be
decomposed as
\begin{equation}
\mathbf{I} = a_{i} \Phi_{i} + \bf{R}
\label{eq:mp0} \end{equation}
where $\bf{R}$ is the residual image (``Pursuit'' step).
We then repeat this 2-step process on the residual (that is, with $\mathbf{I} \leftarrow \bf{R}$)
until some stopping criterion is met.
Note also that the norm of the filters has no influence in this algorithm
on the matching step or on the reconstruction error.
For simplicity and without loss of generality,
we will thereafter set the norm of the filters to $1$: $\forall j, \| \Phi_j \| =1$ (that is, that the spectral energy sums to 1).
Globally, this procedure gives us a sequential algorithm for reconstructing the signal
using the list of sources (filters with coefficients), which greedily optimizes the $\ell_0$ pseudo-norm
(i.e., achieves a relatively sparse representation given the stopping criterion).
The procedure is known as the Matching Pursuit (MP) algorithm~\citep{Mallat93},
which has been shown to generate good approximations for natural images~\citep{Perrinet03ieee,Perrinet10shl}.

We have included two minor improvements over this method:
First, we took advantage of the response of the filters as complex numbers.
As stated above, the modulus gives a response independent of the phase of the filter,
and this value was used to estimate the best match of the residual image
with the possible dictionary of filters (Matching step).
Then, the phase was extracted as the argument of the corresponding coefficient
and used to feed back onto the image in the Pursuit step.
This modification allows for a phase-independent detection of edges,
and therefore for a richer set of configurations,
while preserving the precision of the representation.

%alpha MP (COMP not needed we are at the end of learning scale invariance)
Second, we used a ``smooth'' Pursuit step.
In the original form of the Matching Pursuit algorithm,
the projection of the Matching coefficient is fully removed from the image,
which allows for the optimal decrease of the energy of the residual
and allows for the quickest convergence of the algorithm
with respect to the $\ell_0$ pseudo-norm
(i.e., it rapidly achieves a sparse reconstruction with low error).
However, this efficiency comes at a cost,
because the algorithm may result in non-optimal representations
due to choosing edges sequentially and not globally.
This is often a problem when edges are aligned (e.g. on a smooth contour),
as the different parts will be removed independently, potentially leading
to a residual with gaps in the line.
Our goal here is not necessarily to get the fastest decrease of energy,
but rather to provide with the best representation of edges along contours.
We therefore used a more conservative approach,
removing only a fraction (denoted by $\alpha$)
of the energy at each pursuit step (for MP, $\alpha=1$).
Note that in that case, Equation~\ref{eq:error} has to be modified to account for the $\alpha$ parameter:
\begin{equation}%
E_N =  1 - \alpha\cdot(2-\alpha)\cdot \sum_{1\leq k\leq N} a_{k}^2 / \| \image \| ^2%
\label{eq:error_alpha}%
\end{equation}%
We found that $\alpha=0.8$ was a good compromise between rapidity and smoothness.
One consequence of using $\alpha<1$ is that, when removing energy along contours,
edges can overlap; even so, the correlation is invariably reduced.
Higher and smaller values of $\alpha$ were also tested,
and gave representation results similar to those presented here.

In summary, the whole coding algorithm is given by the following nested loops
in pseudo-code:
\begin{enumerate}
\item draw a signal $\mathbf{I}$ from the database; its energy is $E = \| \mathbf{I} \|^2$,
\item initialize sparse vector $\mathbf{s}$ to zero and linear coefficients $\forall j, {a}_j=<\mathbf{I}, \Phi_j >$,
\item while the residual energy $E = \| \mathbf{I} \|^2$ is above a given threshold do:
\begin{enumerate}
\item select the best match: $i = \mbox{ArgMax}_{j} | {a}_j |$, where $| \cdot |$ denotes the modulus,
\item increment the sparse coefficient: $s_{i} = s_{i} + \alpha \cdot {a}_{i}$,
\item update residual image: $ \mathbf{I} \leftarrow \mathbf{I} - \alpha \cdot a_{i} \cdot \Phi_{i} $,
\item update residual coefficients: $\forall j, {a}_j \leftarrow {a}_j - \alpha \cdot a_{i} <\Phi_{i} , \Phi_j > $,
\end{enumerate}
\item the final set of non-zero values of the sparse representation vector
$\mathbf{s}$ constitutes the list of edges representing the
image as the list of couples $\pi_i = (i, s_{i})$, where $i$ represents an edge occurrence
as represented by its position, orientation and scale and $s_i$ the complex-valued sparse coefficient.
\end{enumerate}
%: TODO say we have lots of controls:	%more N, more \theta, smooth vs hard
This class of algorithms gives a generic and efficient representation of edges,
as illustrated by the example in Figure~\ref{fig:SparseLets}-A.
We also verified that the dictionary used here is better adapted
to the extraction of edges than Gabors~\citep{Fischer07}.
The performance of the algorithm can be measured quantitatively
by reconstructing the image from the list of extracted edges.
All simulations were performed using \verb+Python+ (version 2.7.8)
with packages \verb+NumPy+ (version 1.8.1) and \verb+SciPy+ (version 0.14.0)~\citep{Oliphant07}
on a cluster of Linux computing nodes.
Visualization was performed using \verb+Matplotlib+ (version 1.3.1)~\citep{Hunter07}.
\footnote{These python scripts are available at \url{https://github.com/bicv/SparseEdges} and documented at \url{https://pythonhosted.org/SparseEdges}.}
%Python	2.7.8 (default, Jul 2 2014, 10:14:46) [GCC 4.2.1 Compatible Apple LLVM 5.1 (clang-503.0.40)]
%IPython	2.1.0
%OS	posix [darwin]
%numpy	1.8.1
%scipy	0.14.0
%matplotlib	1.3.1
\subsection{Efficiency of the SparseLets framework}
%~~~~~~~~~~~~~~~~~~~~~~~~~~~~~~~~~~~~~~~~~~%
%------------------------------%
%: See Figure~\ref{fig:SparseLets}
\begin{figure}[ht!]%[p!]
\centering{
\begin{tikzpicture}
\draw [anchor=north west] (0, 0.) node {\includegraphics[width=.382\textwidth]{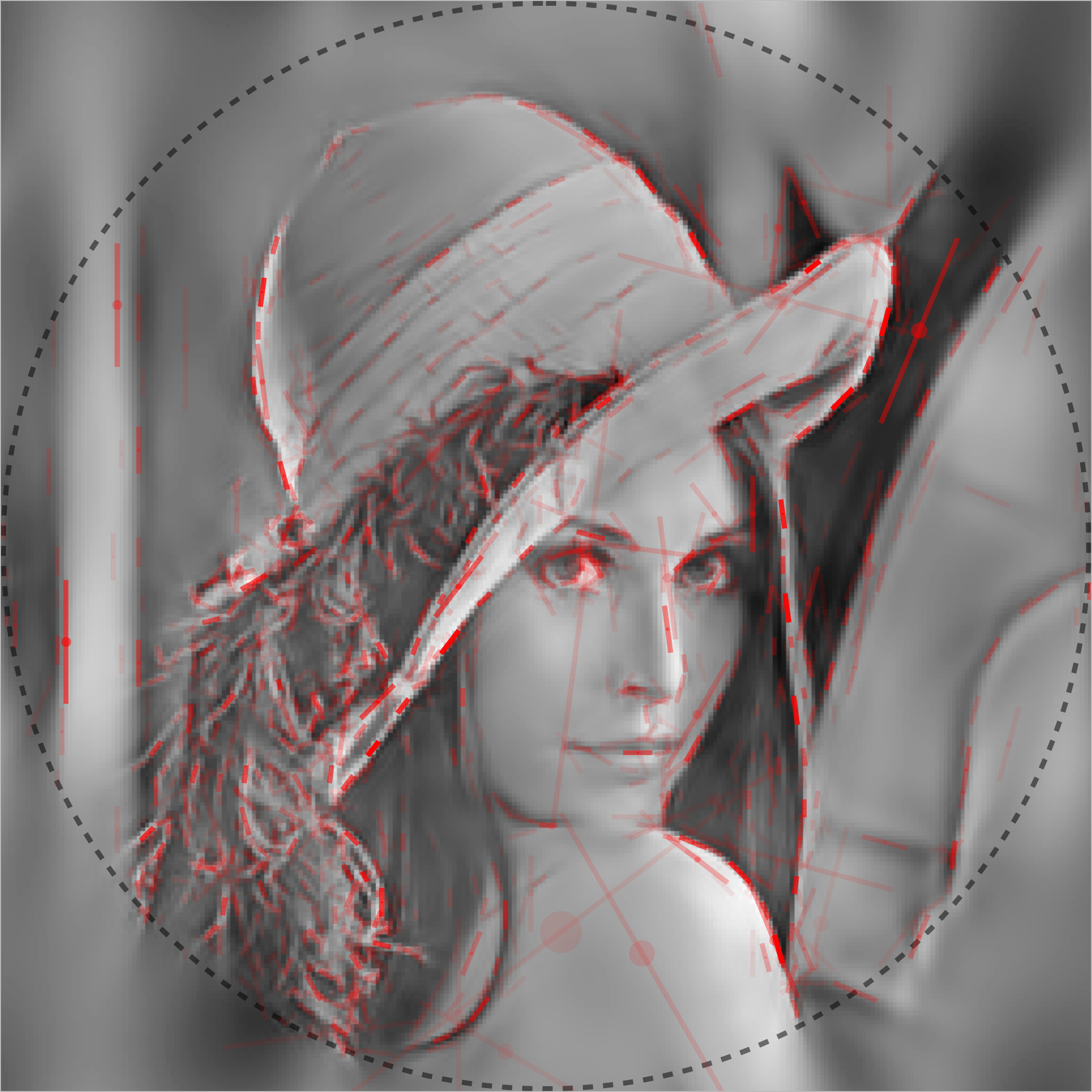}};
%\draw[black, dashed, thick] (.404\textwidth/2, -.404\textwidth/2) circle (.382\textwidth/2);
\draw [anchor=north west] (.382\textwidth, 0.) node {\includegraphics[height=.382\textwidth]{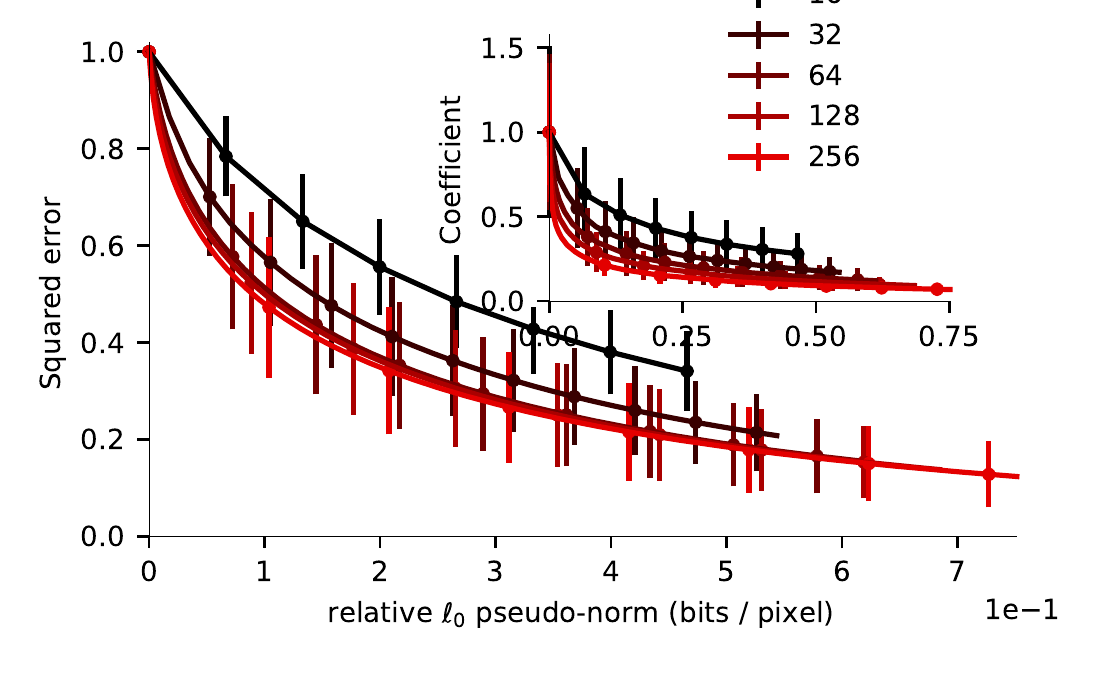}};
\draw (0, 0.) node [above right=0mm] {$\mathsf{(A)}$};
\draw (.382\textwidth, 0.) node [above right=0mm] {$\mathsf{(B)}$};
\end{tikzpicture}}
\caption{ {\bf SparseLets framework.} \textsf{(A)}~ An example reconstructed image with the list of extracted edges overlaid.
As in~\citet{Geisler01}, edges outside the circle are discarded to avoid artifacts.
Parameters for each edge are the position, , orientation, scale (length of bar) and scalar amplitude (transparency) with the phase (hue).
We controlled the quality of the reconstruction from the edge information such
that the residual energy is less than $3\%$ over the whole set of images,
a criterion met on average when identifying $2048$ edges per image for images of size $256\times 256$ (that is, a relative sparseness of $\approx 0.01\%$ of activated coefficients).
\textsf{(B)}~Efficiency for different image sizes as measured by the decrease of the residual's energy as a function of the coding cost
%ratio of active coefficients 
(relative $\ell_0$ pseudo-norm).
\textsf{(B,~inset)}~%Similarly to Figure~\ref{fig:sparsenet}, the decrease of coefficients follows a power-law.
%We compared the strength of this power-law on a image-per-image basis (as shown by the error bars) as a function of the image size.
This shows that as the size of images increases, sparseness increases,
validating quantitatively our intuition on the sparse positioning of objects in natural images.
Note, that the improvement is non significant for a size superior to $128$.
The SparseLets framework thus shows that sparse models can be extended to full-scale natural images, and that increasing the size improves sparse models by a degree of order (compare a size of $16$ with that of $256$).
\label{fig:SparseLets}}%
\end{figure}%
%------------------------------%

% vanilla version RMSE always achieves 5% / power-law
Figure~\ref{fig:SparseLets}-A, shows the list of edges extracted on a sample image. It fits qualitatively well with a rough sketch of the image.
To evaluate the algorithm quantitatively,
we measured the ratio of extracted energy in the images as a function of the number of edges
on a database of 600 natural images  of size $256\times 256$\footnote{This database is publicly available at \url{http://cbcl.mit.edu/software-datasets/serre/SerreOlivaPoggioPNAS07}.}, see Figure~\ref{fig:SparseLets}-B.
Measuring the ratio of extracted energy in the images, $N=2048$ edges were
enough to extract an average of approximately $97\%$ of the energy of images in the database. %
%A number $N=1024$ of edges were
%enough to extract an average of $95\%$ of the energy of $256\times 256$
%images on the set of images. %
To compare the strength of the sparseness in these full-scale images
compared to the image patches discussed above (see Section~\ref{sec:sparsenet}),
we measured the sparseness values obtained in images of different sizes.
To be comparable, we measured the efficiency with respect to the relative $\ell_0$ pseudo-norm in bits per unit of image surface (pixel):
This is defined as the ratio of active coefficients times the numbers of bits required to code for each coefficient (that is, $\log_2(M)$, where $M$ is total number of coefficients in the representation) over the size of the image. 
For different image and framework sizes, the lower this ratio, the higher the sparseness.
As shown in Figure~\ref{fig:SparseLets}-B,
we indeed see that sparseness increases relative to an increase in image size.
This reflects the fact that sparseness is not only local (few edges coexist at one place)
but is also spatial (edges are clustered, and most regions are empty).
Such a behavior is also observed in V1 of monkeys
as the size of the stimulation is increased from a stimulation over only the classical receptive field
to 4 times around it~\citep{Vinje00}.

Note that by definition, our representation of edges
is invariant to translations, scalings, and rotations in the plane of the image.
We also performed the same edge extraction where images from the database
were perturbed by adding independent Gaussian noise to each pixel
such that signal-to-noise ratio was halved.
Qualitative results are degraded but qualitatively similar.
In particular, edge extraction in the presence of noise may result in false positives.
Quantitatively, one observes that the representation is slightly less sparse.
This confirms our intuition that sparseness is causally linked to the efficient extraction of edges in the image.

\begin{figure}%[ht!]%[p!]
\centering{%
\includegraphics[width=.618\textwidth]{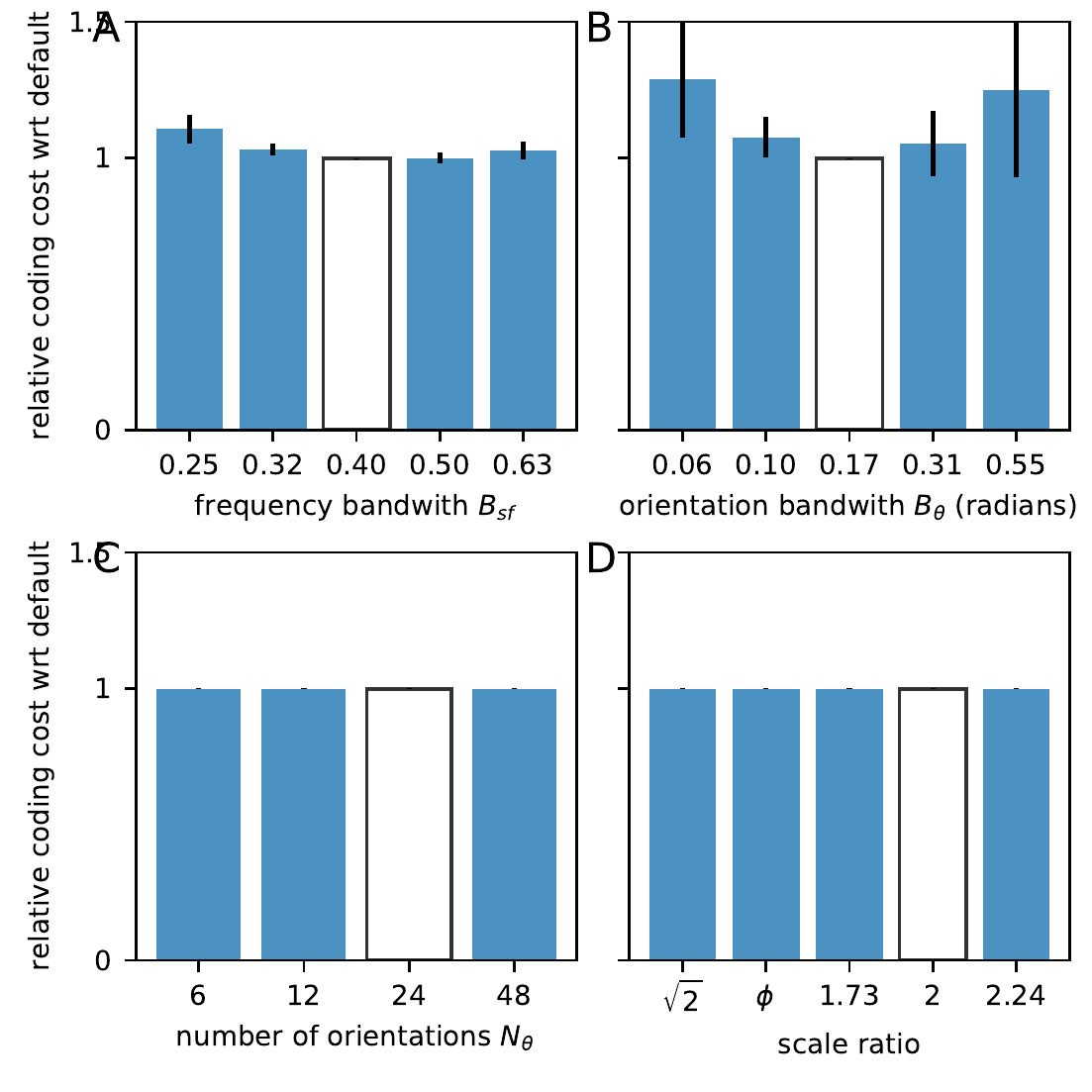}
%\begin{tikzpicture}
%%\draw [anchor=north west] (0, .618\textwidth) node {\includegraphics[width=.5\textwidth,height=.309\textwidth]{efficiency_A}};
%%\draw [anchor=north west] (.5\textwidth, .618\textwidth) node {\includegraphics[width=.5\textwidth,height=.309\textwidth]{efficiency_B}};
%%\draw [anchor=north west] (0, .309\textwidth) node {\includegraphics[width=.5\textwidth,height=.309\textwidth]{efficiency_C}};
%%\draw [anchor=north west] (.5\textwidth, .309\textwidth) node {\includegraphics[width=.5\textwidth,height=.309\textwidth]{efficiency_D}};
%\draw [anchor=south west] (-.2, 0) node {\includegraphics{efficiency}};
%
%\draw (0, \textwidth) node [above right=0mm] {$\mathsf{(A)}$};
%\draw (.5\textwidth, \textwidth) node [above right=0mm] {$\mathsf{(B)}$};
%\draw (0, .5\textwidth) node [above right=0mm] {$\mathsf{(C)}$};
%\draw (.5\textwidth, .5\textwidth) node [above right=0mm] {$\mathsf{(D)}$};
%\end{tikzpicture}
}
\caption{ {\bf Effect of filters' parameters on the efficiency of the {SparseLets} framework } As we tested different parameters for the filters, we measured the gain in efficiency for the algorithm as the ratio of the code length to achieve 85\% of energy extraction relative to that for the default parameters (white bar). The average is computed on the same database of natural images and error bars denote the standard deviation of gain over the database. First, we studied the effect of the bandwidth of filters respectively in the \textsf{(A)}~spatial frequency and \textsf{(B)}~orientation spaces. The minimum is reached for the default parameters: this shows that default parameters provide an optimal compromise between the precision of filters in the frequency and position domains for this database. We may also compare pyramids with different number of filters.  Indeed from Equation~\ref{eq:L0_cost}, efficiency (in bits) is equal to the number of selected filters times the coding cost for the address of each edge in the pyramid.
We plot here the average gain in efficiency which shows an optimal compromise respectively for respectively \textsf{(C)}~the number of orientations and \textsf{(D)}~the number of spatial frequencies (scales). Note first that with more than 12 directions, the gain remains stable. Note also that a dyadic scale ratio (that is of 2) is efficient but that other solutions ---such as using the golden section $\phi$--- prove to be significantly more efficient, though the average gain is relatively small (inferior to 5\%).
\label{fig:efficiency}}%
\end{figure}

To examine the robustness of the framework and of sparse models in general,
we examined how results changed when changing  parameters for the algorithm.
In particular, we investigated the effect of filter parameters $B_{f}$ and $B_{\theta}$.
We also investigated how the over-completeness factor
could influence the result. We manipulated the number of discretization steps along the spatial frequency axis $N_f$ (that is, the number of layers in the pyramid) and orientation axis $N_\theta$.
Results are summarized in Figure~\ref{fig:efficiency}
and show that an optimal efficiency is achieved for certain values of these parameters.
These optimal values are in the order of what is found for the range of selectivities observed in V1.
Note that these values may change across categories.
Further experiments should provide
with an adaptation mechanism
to allow finding the  best parameters in an unsupervised manner. %/ compromise

%\showthe\columnwidth
%------------------------------%
%: See Figure~\ref{fig:efficiency}
%------------------------------%

These particular results illustrate the potential of sparse models in computer vision.
Indeed, one main advantage of these methods is to explicitly represent edges.
A direct application of sparse models is the ability of the representation to reconstruct these images and therefore to use it for compression~\citep{Perrinet03ieee}.
Other possible applications are image filtering or edge manipulation for texture synthesis or denoising~\citep{Portilla00}.
Recent advances have shown that such representations could be used for the classification of natural images (see chapter \verb+013_theriault+ or for instance~\citep{PerrinetBednar15}) or of medical images of emphysema~\citep{Nava13}.
Classification was also used in a sparse model
for the quantification of artistic style through sparse coding analysis
in the drawings of Pieter Bruegel the Elder~\citep{Hughes10}.
These examples illustrate the different applications of sparse representations and in the following we will illustrate some potential perspectives to further improve their representation efficiency.
\section{SparseEdges: introducing prior information}
%-------------------------------------------------%
\label{sec:sparseedges}
\subsection{Using the prior in first-order statistics of edges}
%~~~~~~~~~~~~~~~~~~~~~~~~~~~~~~~~~~~~~~~~~~~~~~~~~~~~~~~~~~~~~%
\label{sec:laughlin}
% neural code and equalization: observations
In natural images, it has been observed that edges follow some statistical regularities that may be used by the visual system.
We will first focus on the most obvious regularity which consists in the anisotropic distribution of orientations in natural images (see chapter \verb+003_holly_gerhard+ for another qualitative characterization of this anisotropy).
Indeed, it has been observed that orientations corresponding to cardinals (that is, to verticals and horizontals) are more likely than other orientations~\citep{Ganguli10,Girshick11}.
This is due to the fact that our point of view is most likely pointing toward the horizon while we stand  upright. In addition, gravity shaped our surrounding world around horizontals (mainly the ground) and verticals (such as trees or buildings). Psychophysically, this prior knowledge gives rise to the oblique effect~\citep{Keil00}.
This is even more striking in images of human scenes (such as a street, or inside a building) as humans mainly build their environment (houses, furnitures) around these cardinal axes. However, we assumed in the cost defined above (see~\seeEq{efficiency_cost}) that each coefficient is independently distributed.

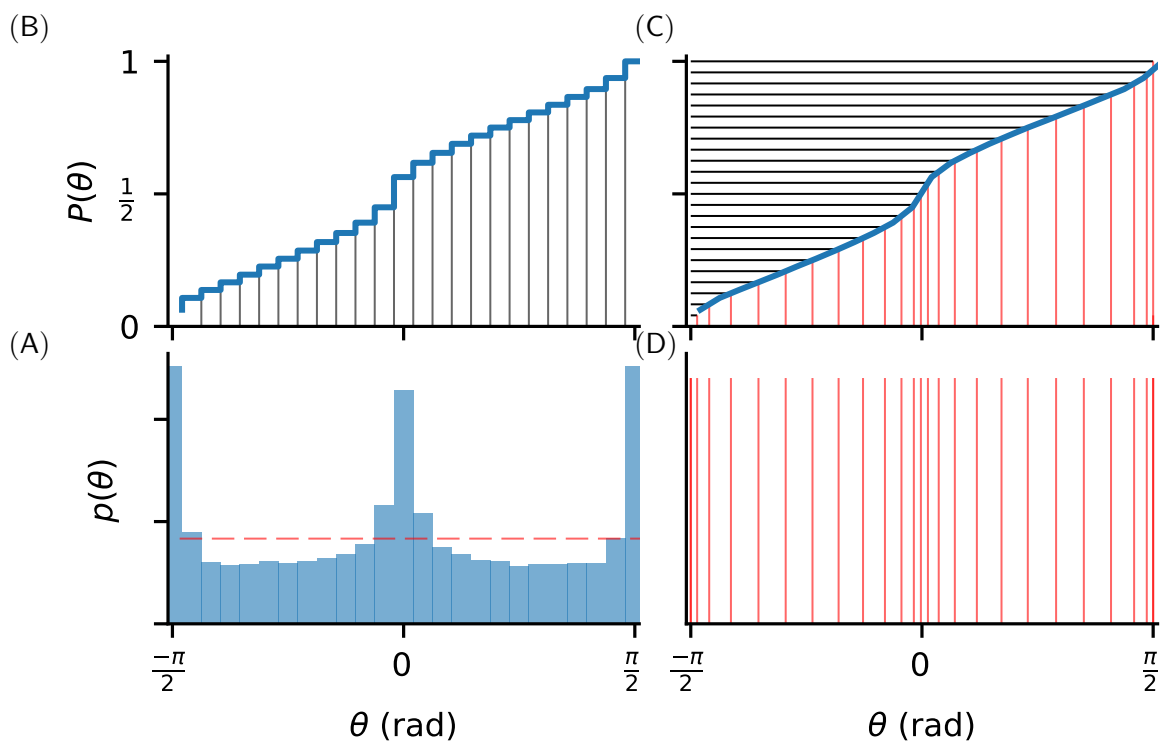
\begin{figure}[ht!]%[p!]
\centering{%
\begin{tikzpicture}
\draw [anchor=north west] (0, .618\textwidth) node {\includegraphics[width=1.\columnwidth]{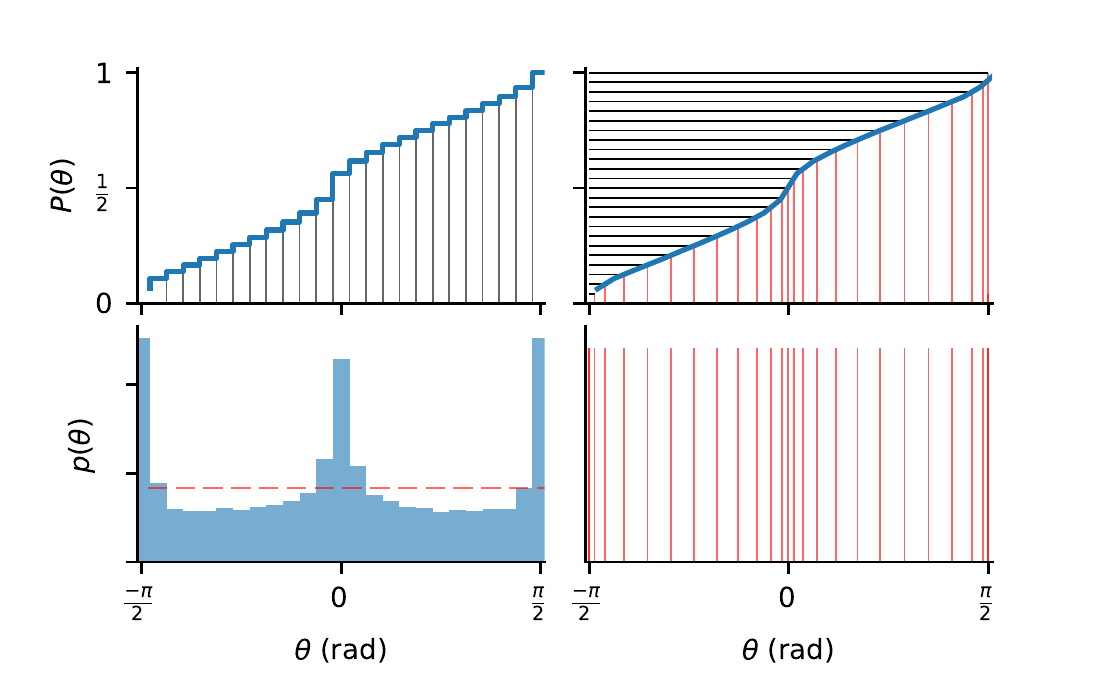}};

\draw (.0\textwidth, .55\textwidth) node [above right=0mm] {$\mathsf{(B)}$};
\draw (.49\textwidth, .55\textwidth) node [above right=0mm] {$\mathsf{(C)}$};
\draw (.0\textwidth, .3\textwidth) node [above right=0mm] {$\mathsf{(A)}$};
\draw (.49\textwidth, .3\textwidth) node [above right=0mm] {$\mathsf{(D)}$};
\end{tikzpicture}}
\caption{ {\bf Histogram equalization} From the edges extracted in the images from the natural scenes database, we computed sequentially (clockwise, from the bottom left): \textsf{(A)}~the histogram  and \textsf{(B)}~cumulative histogram of edge orientations. This shows that as was reported previously (see for instance~\citep{Girshick11}), cardinals axis are over-represented. This represents a relative inefficiency as the representation in the SparseLets framework represents a priori orientations in an uniform manner. A neuromorphic solution is to use histogram equalization, as was first shown in the fly's compound eye by~\citep{Laughlin81}. \textsf{(C)}~We draw a uniform set of scores on the y-axis of the cumulative function (black horizontal lines), for which we select the corresponding orientations (red vertical lines). Note that by convention these are wrapped up to fit in the $( -\pi/2, \pi/2]$ range. \textsf{(D)}~This new set of orientations is defined such that they are \emph{a priori} selected uniformly. Such transformation was found to well describe a range of psychological observations~\citep{Ganguli10} and we will now apply it to our framework.
\label{fig:laughlin}}%
\end{figure}

It is believed that an homeostasis mechanism allows one to optimize this cost knowing this prior information~\citep{Laughlin81,Perrinet10shl}.
Basically, the solution is to put more filters where there are more orientations~\citep{Ganguli10} such that coefficients are uniformly distributed.
In fact, since neural activity in the assembly actually represents the sparse coefficients, we may understand the role of homeostasis as maximizing the average representation cost $\mathcal{C}( \coef | \dico)$. This is equivalent to saying that homeostasis should act such that at any time, and invariantly to the selectivity of features in the dictionary, the probability of selecting one feature is uniform across the dictionary.
This optimal uniformity may be achieved in all generality by using an equalization of the histogram~\citep{Atick92}. This method may be easily derived if we know the probability distribution function $dP_{i}$ of variable $a_i$  (see~\seeFig{laughlin}-A) by choosing a non-linearity as the cumulative distribution function (see~\seeFig{laughlin}-B) transforming any observed variable $\bar{a}_i$ into:
\begin{equation}
z_i(\bar{a}_i)= P_{i}(a_i \leq \bar{a}_i) = \int_{-\infty}^{\bar{a}_i} dP_{i}(a_i)%
\label{eq:laughlin}%
\end{equation}
This is equivalent to the change of variables which transforms the sparse vector $\coef$ to a variable with uniform probability distribution function in $[0, 1]^M$ (see~\seeFig{laughlin}-C).
This equalization process has been observed in the neural activity of a variety of species and is, for instance, perfectly illustrated  in the compound eye of the fly's neural response to different levels of contrast~\citep{Laughlin81}. It may evolve dynamically to slowly adapt to varying changes, for instance to luminance or contrast values, such as when the light diminishes at twilight.
Then, we use these point non-linearities $z_i$ to sample orientation space in an optimal fashion (see~\seeFig{laughlin}-D). %
%------------------------------%
%: See Figure~\ref{fig:laughlin}
%------------------------------%

%------------------------------%
%: See Figure~\ref{fig:firstorder}
\begin{figure}%[ht!]%[p!]
\centering{
\begin{tikzpicture}
\node [anchor=south west] (a) at (0, 0.) {\includegraphics[width=.333\textwidth]{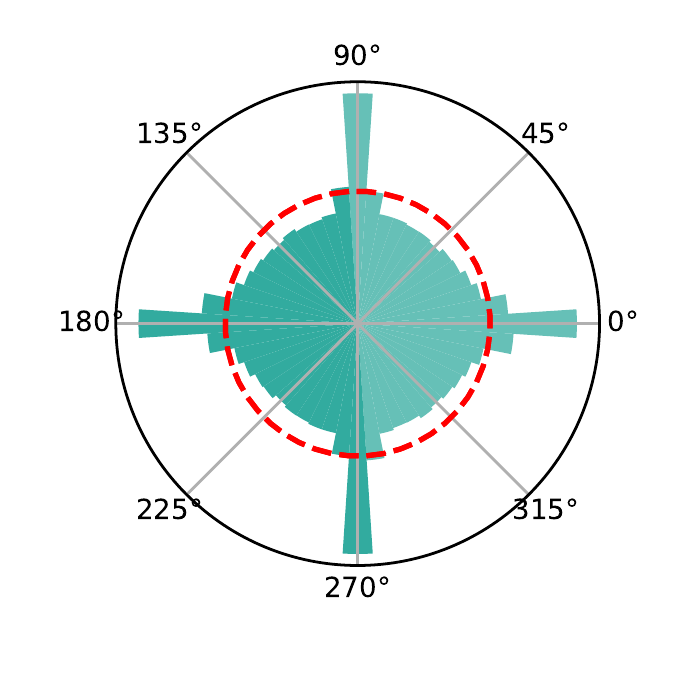}};
\node [anchor=south west] (a) at (.333\textwidth, 0.) {\includegraphics[width=.333\textwidth]{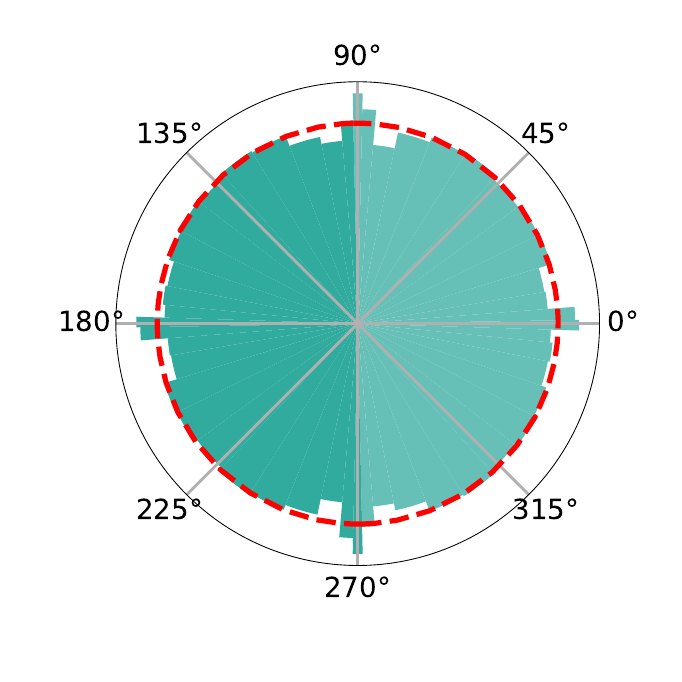}};
\node [anchor=south west] (b) at (.666\textwidth, 0.) {\includegraphics[width=.333\textwidth]{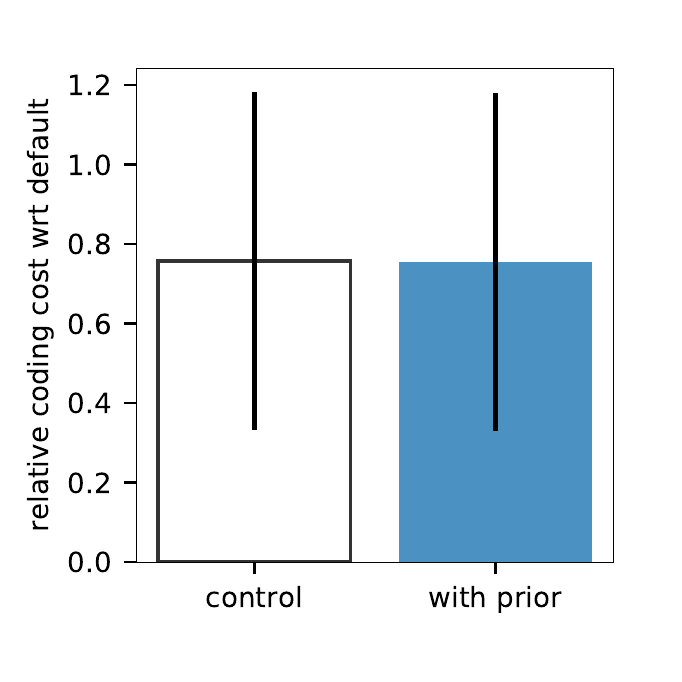}};

\draw [anchor=south west] (0, .333\textwidth) node {$\mathsf{(A)}$};
\draw [anchor=south west] (.333\textwidth, .333\textwidth) node {$\mathsf{(B)}$};
\draw [anchor=south west] (.666\textwidth, .333\textwidth) node {$\mathsf{(C
)}$};
\end{tikzpicture}}
\caption{ {\bf Including prior knowledge on the orientation of edges in natural images.}  \textsf{(A)}~Statistics of orientations in the database of natural images as shown by a polar bar histogram (the surface of wedges is proportional to occurrences). As for Figure~\ref{fig:laughlin}-A, this shows that orientations around cardinals are more likely than others (the dotted red line shows the average occurrence rate). \textsf{(B)}~ We show the histogram with the new set of orientations that has been used. Each bin is selected approximately uniformly. Note the variable width of bins. 
\textsf{(C)}~We compare the efficiency of the modified algorithm where the sampling is optimized thanks to histogram equalization described in Figure~\ref{fig:laughlin} as the average residual energy with respect to the number of edges. %  (in the inset, we similarly plot sorted coefficients).
This shows that  introducing a prior information on the distribution of orientations in the algorithm may also introduce a slight but insignificant improvement in the sparseness. % (of the order of $5\%$) and therefore in the efficiency of the representation.
\label{fig:firstorder}}%
\end{figure}%
%------------------------------%
This simple non-parametric homeostatic method is applicable to the SparseLets algorithm by simply using the transformed sampling of the orientation space.
It is important to note that the MP algorithm is non-linear and the choice of one element at any step may influence the rest of the choices.
%Results show that by using this prior knowledge, one achieves a better coding of images.
%Indeed, by looking at the qualitative results, we could see that edges that are more common are more finely detected.
In particular, while orientations around cardinals are more prevalent in natural images (see Figure~\ref{fig:firstorder}-A), the output histogram of detected edges is uniform (see Figure~\ref{fig:firstorder}-B). %and these were detected with a better precision.
To quantify the gain in efficiency, we measured the residual energy in the SparseLets framework with or without including this prior knowledge. Results show that for a similar number of extracted edges, residual energy is not significantly changed (see Figure~\ref{fig:firstorder}-C).
This is again due to the exponential convergence of the squared error~\citep[p.~422]{Mallat98} on the space spanned by the representation basis.
As the tiling of the Fourier space by the set of filters is complete, 
one is assured of the convergence of the representation in both cases.
%This indicates that in general, representation is more sparse and therefore that it better reflects the physical representation of the synthesis of images.
%Moreover, this measures only accounts for the residual energy as a function of the number of coefficients.
However thanks to the use of first-order statistics, the orientation of edges are distributed such as to maximize the entropy, further improving the efficiency of the representation.

This novel improvement to the SparseLets algorithm illustrates the flexibility of the Matching Pursuit framework. %, and one would expect that the residuals that are extracted are of lower energy and therefore that the decrease is slower.
This proves that by introducing the prior on first-order statistics, one improves the efficiency of the model for this class of natural images.
Of course, this gain is only valid for natural images and would disappear for images where cardinals would not dominate. This is the case for images of close-ups (microscopy) or where gravity is not prevalent such as aerial views.
Moreover, this is obviously just a first step as there is more information from natural images that could be taken into account.

% a note on frquentist probabilities computsd feom natural omages and bayesian statistics computzd to make infernces on beliefso
% todo check biblio of david field :-
% - 87 stats
% 93  path paradigm 94 what is the goal
% 99 log gabor
%Field, D. J., Hayes, A. & Hess, R. F. 1993 Contour Integration by the Human Visualsystem: Evidence for a Local "Association Field". Vision Research  33, 173-193. Field, D. J., Hayes, A. & Hess, R. F. 1997 The role of phase and contrast polarity in contour integration. Investigative Ophthalmology & Visual Science 38, 4643-4643.

\subsection{Using the prior statistics of edge co-occurences}
%~~~~~~~~~~~~~~~~~~~~~~~~~~~~~~~~~~~~~~~~~~~~~~~~~~~~~~~~~~~~~~~~~~~%
%observations -Geisler01 Sigman01- quantifying assofield
A natural extension of the previous result is to study the co-occurrences of edges in natural images.
Indeed, images are not simply built from independent edges at arbitrary positions and orientations but tend to be organized along smooth contours that follow for instance the shape of objects.
In particular, it has been shown that contours are more likely to be organized along co-circular shapes~\citep{Sigman01}.
This reflects the fact that in nature, round objects are more likely to appear than random shapes.
Such a statistical property of images seems to be used by the visual system as it is observed that edge information is integrated on a local "association field" favoring co-linear or co-circular edges (see chapter \verb+013_theriault+ section 5 for more details and a mathematical description).
In V1 for instance, neurons coding for neighboring positions are organized in a similar fashion. We have previously seen that statistically, neurons coding for collinear edges seem to be anatomically connected~\citep{Bosking97,Hunt11} while rare events (such as perpendicular occurrences) are functionally inhibited~\citep{Hunt11}.

%------------------------------%
%: See Figure~\ref{fig:secondorder}
\begin{figure}%[ht!]%[p!]
\centering{
\begin{tikzpicture}[font=\sffamily]
% (C) edge schematics
\begin{scope}[scale=.75, yshift=2.cm, xshift=-1.cm]
 %\fill[white,fill opacity=.9] (0, 0) rectangle (9, 9);
% \tkzInit[xmax=12, ymax=6]

 %----------------------------------------------------------
 % Defining coordinates
 %----------------------------------------------------------

 \tkzDefPoint(2.95, 2.25){A}
 \tkzDefPoint(8, 5.25){B}
 \tkzLabelPoints[above left](B,A)
 \tkzDefPoint(7.5, 2.25){C}

 % draw red dots at the center of edges
 \tkzDrawPoints[size=10, color=red, fill=red](A,B)

 %----------------------------------------------------------
 % Drawing the lines and segments
 %----------------------------------------------------------
 \tkzDrawLine[color=red,line width=2pt, add=-1.15 and -.15 ](C,A)
 \tkzDrawLine[color=red,line width=2pt, add=-1.3 and -.3 ](C,B)

 \tkzDrawLine(A,B)
 \tkzDrawLines[dashed](A,C B,C)

 % drawing arcs for angles
 \tkzMarkAngle[size=2.5,mkpos=.2](C,A,B)
 \tkzLabelAngle[pos=3.5,circle](C,A,B){$\mathsf{\phi}$}

 \tkzDefPointWith[linear,K=1.5](A,C)
 \tkzGetPoint{D}
 \tkzDefPointWith[linear,K=.75](B,C)
 \tkzGetPoint{E}
 \tkzMarkAngle[size=1,mkpos=.2](D,C,E)
 \tkzLabelAngle[pos=1.75,circle](D,C,E){$\mathsf{\theta}$}

 %----------------------------------------------------------
 % Drawing normals
 %----------------------------------------------------------

 \tkzDefLine[perpendicular=through A, K=-.75](C,A)
 \tkzGetPoint{a1}
 \tkzDefLine[perpendicular=through B, K=1.5](C,B)
 \tkzGetPoint{b1}
 \tkzInterLL(a1,A)(b1,B) \tkzGetPoint{H}
 \tkzMarkRightAngle[size=.5](H,A,C)
 \tkzMarkRightAngle[size=.5](H,B,C)
 \tkzDrawLines[dashed,dash phase=1.5pt](a1,A)
 \tkzDrawLines[dashed,dash phase=0.5pt](b1,B)

 %----------------------------------------------------------
 % Drawing mediator and psi line
 %----------------------------------------------------------
% \tkzDefLine[mediator](A,B) \tkzGetPoints{m1}{M}
 \tkzDefMidPoint(A,B)
 \tkzGetPoint{M}
 \tkzDefLine[perpendicular=through M, K=.4](A,B)
 \tkzGetPoint{m1}
 \tkzMarkRightAngle[size=.5](B,M,m1)

 \tkzFillAngle[size=1.4,fill=blue!40](m1,M,H)
 \tkzLabelAngle[pos=2,circle](m1,M,H){$\mathsf{\psi}$}%= \phi -\theta/2
 \tkzDrawLines[](m1,M M,H)
 \end{scope}
\node [anchor=south west] (chevrons) at (.4\textwidth, 0.) {\includegraphics[width=.6\columnwidth]{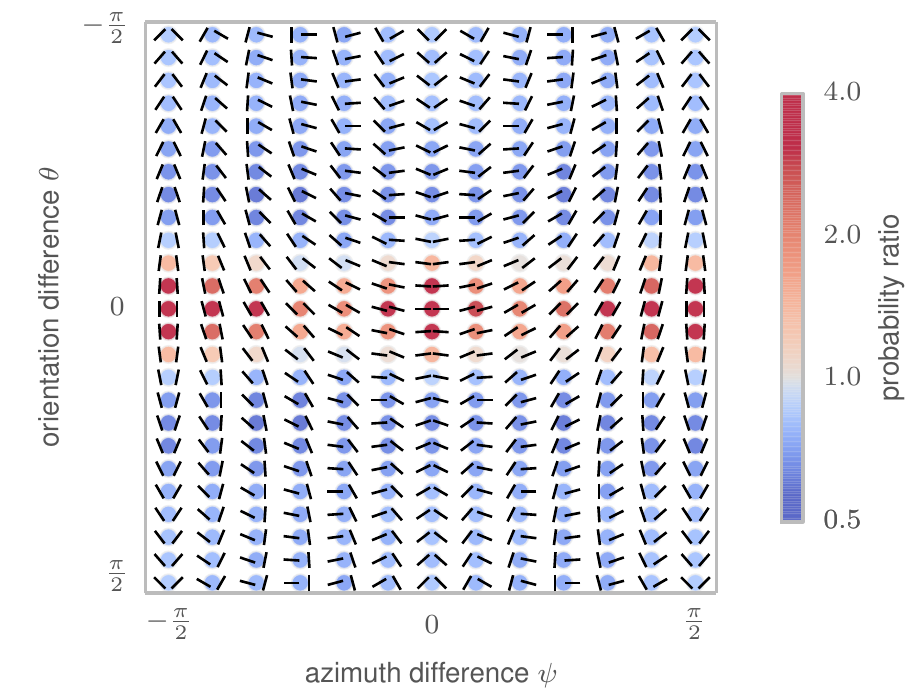}};
\draw [anchor=south west] (0, .43\textwidth) node {$\mathsf{(A)}$};
\draw [anchor=south west] (.4\textwidth, .43\textwidth) node {$\mathsf{(B)}$};
\end{tikzpicture}
}
\vspace*{-1.cm}
\caption{{\bf Statistics of edge co-occurences}
 \textsf{(A)}~The relationship between a pair of edges can be quantified
in terms of the difference between their orientations $\theta$,
the ratio of scale $\sigma$ relative to the reference edge,
the distance $d=\|\vec{AB}\|$ between their centers,
and the difference of azimuth (angular location) $\phi$
of the second edge relative to the reference edge.
Additionally, we define $\psi=\phi - \theta/2$ as it is
symmetric with respect to the choice of the reference edge,
%todo psi was defined in hunt 11 figure 3b as the half of the angle in the cotangent circle
in particular, $\psi=0$ for co-circular edges.
\textsf{(B)}~The probability distribution function $p(\psi, \theta)$
represents the distribution of the different geometrical arrangements of edges' angles,
which we call a ``chevron map''.
We show here the histogram for natural images,
illustrating the preference for co-linear edge configurations.
For each chevron configuration, deeper and deeper red circles
indicate configurations that are more and more likely
with respect to a uniform prior, % (all configurations equally likely),
with an average maximum of about $4$ times more likely,
and deeper and deeper blue circles indicate configurations less likely
than a flat prior (with a minimum of about $0.7$ times as likely).
Conveniently, this ``chevron map'' shows in one graph that natural images
have on average a preference for co-linear and parallel angles
(the horizontal middle axis), %as seen in Figure~\ref{fig:model}-D,
along with a slight preference for co-circular configurations (middle vertical axis).
\label{fig:secondorder}}%
\end{figure}%
%------------------------------%

\begin{figure}[ht!]%[p!]
\centering{
\begin{tikzpicture}
\node [anchor=north west] (a) at (0, 0.) {\includegraphics[width=.33\textwidth]{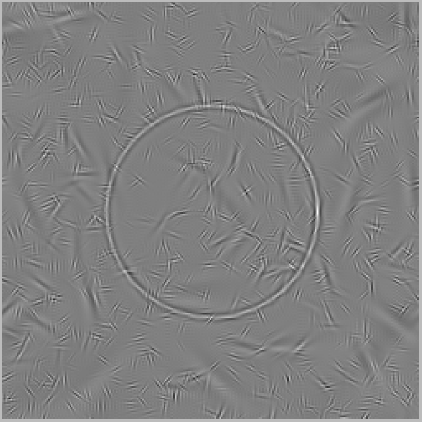}};
\node [anchor=north west] (a) at (.33\textwidth, 0.) {\includegraphics[width=.33\textwidth]{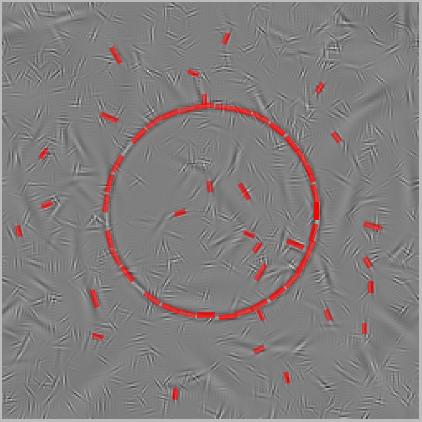}};
\node [anchor=north west] (b) at (.66\textwidth, 0.) {\includegraphics[height=.33\textwidth]{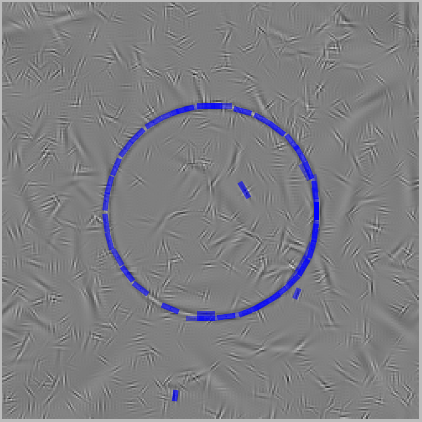}};

\draw [anchor=south west] (0, 0) node {$\mathsf{(A)}$};
\draw [anchor=south west] (.33\textwidth, 0) node {$\mathsf{(B)}$};
\draw [anchor=south west] (.66\textwidth, 0) node {$\mathsf{(C)}$};
\end{tikzpicture}}

\caption{{\bf Application to rapid contour segmentation}
We applied the original sparse edge framework to \textsf{(A)}~the synthetic image of a circle embedded in noise. This noise is synthesized by edges at random positions, orientations and scales with a similar first-order statistics as natural scenes. % in~\citep[Fig.~7A]{Geisler01}. 
\textsf{(B)} We overlay in red the set of edges which were detected by the SparseLets framework.
\textsf{(C)}  Then, we have introduced second-order information in the evaluation of the probability in the sparse edges framework (with $\eta=0.15$). This modified the sequence of extracted edges as shown in blue. There is a clear match of the  edge extraction with the circle, as would be predicted by psychophysical results of hand segmentation of contours. % (figure modified from~\citep[Fig.~7B]{Geisler01}). 
This shows that second-order information as introduced in this feed-forward chain  may be sufficient to account for contour grouping and may not necessitate a recursive chain rule such as implemented in~\citet{Geisler01}.
\label{fig:segmentation}}%
\end{figure}
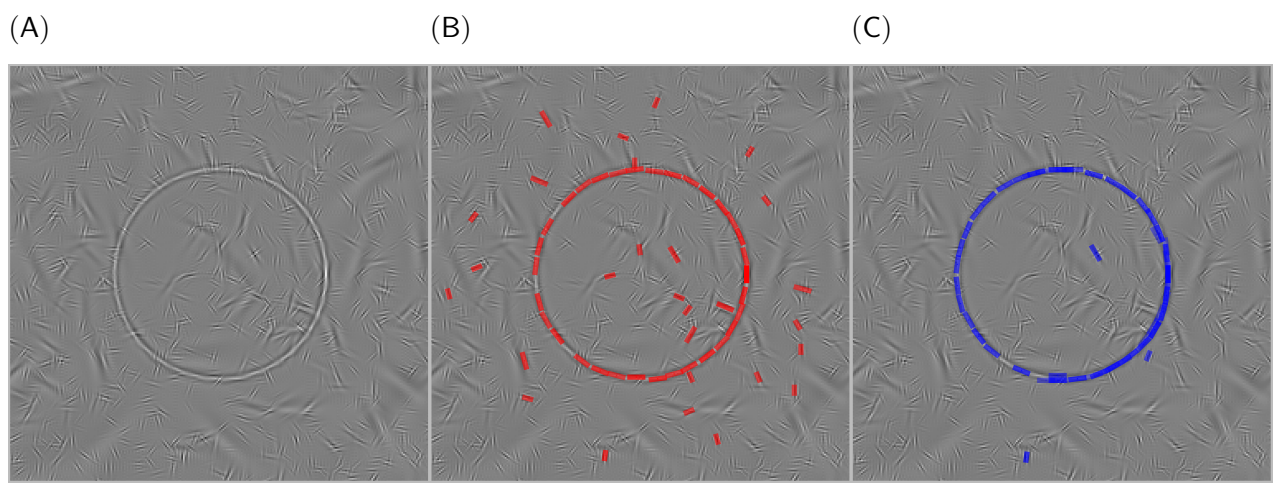

Using the probabilistic formulation of the edge extraction process (see Section~\ref{sec:sparsenet}), one can also apply this prior probability to the choice mechanism (Matching) of the Matching Pursuit algorithm.
Indeed at any step of the edge extraction process, one can include the knowledge gained by the extraction of previous edges, that is, the set  $\mathcal{I} =\{\pi_i\}$ of extracted edges, to refine the log-likelihood of a new possible edge $\pi_\ast=(\ast, a_\ast)$ (where $\ast$ corresponds to the address of the chosen filter, and therefore to its position, orientation and scale).
Knowing the probability of co-occurences $p ( \pi_\ast | \pi_i)$ from the statistics observed in natural images (see Figure~\ref{fig:secondorder}), we deduce that the cost is now at any coding step (where $\image$ is the residual image ---see Equation~\ref{eq:mp0}):
\begin{eqnarray}%
\mathcal{C}(\pi_\ast | \image, \mathcal{I}) = \frac{1}{2\sigma_n^2} \| \image - a_\ast \Phi_\ast \|^2  - \eta \sum_{i\in \mathcal{I}} \| a_i \| \cdot \log p ( \pi_\ast | \pi_i)%
\label{eq:efficiency_cooc}%
\end{eqnarray}%
where $\eta$ quantifies the strength of this prediction. 
Basically, this shows that, similarly to the association field proposed by~\citep{Grossberg84} which was subsequently observed in cortical neurons~\citep{vonderHeydt84} and applied by~\citep{Field93},
we facilitate the activity of edges knowing the list of edges that were already extracted.
This comes as a complementary local interaction to the inhibitory local interaction implemented in the Pursuit step (see Equation~\ref{eq:mp0})
and provides a quantitative algorithm to the heuristics proposed in~\citep{Fischer07}.
Note that though this model is purely sequential and feed-forward, this results possibly in a "chain rule" as when edges along a contour are extracted,
this activity is facilitated along it as long as the image of this contour exists in the residual image.
%In particular, a closed contour will be enhanced, such as been reported psychophysically~\citep{Kovacs93}.
Such a ``chain rule'' is similar to what was used to model psychophysical performance~\citep{Geisler01} or to filter curves in images~\citep{August01}.
Our novel implementation provides with a rapid and efficient solution that we illustrate here on a segmentation problem (see Figure~\ref{fig:segmentation}).

%------------------------------%
%Sum product rule allow to rate the probability of each edge individually : classification on different categories of objects
Indeed, sparse models have been shown to foster numerous applications in computer vision.
Among these are algorithms for segmentation in images~\citep{Spratling13} or for classification~\citep{Spratling13bicy,Dumoulin14}.
We may use the previous application of our algorithm to evaluate the probability of edges belonging to the same contour.
%For that, once all edges were extracted, we  used the above rightmost part of the equation to form the following chain rule to evaluate if a given edge belongs to a contour $\mathcal{S}$:
%\begin{eqnarray}%
%p ( i \in \mathcal{S} ) = \sum_j  p ( i | j) \cdot p ( j \in \mathcal{S} )%
%\label{eq:seg_cooc}%
%\end{eqnarray}%
%Such an equation is solved iteratively by starting with a uniform distribution for the distribution $ p ( j \in \mathcal{S} )$.
We show in Figure~\ref{fig:segmentation} the application of such a formula (in panel C versus classical sparse edge extraction in panel B) on a synthetic image of a circle embedded in noise (panel A). %used by~\citep[Fig.~7A]{Geisler01}. 
It shows that, while some edges from the background are extracted in the plain SparseLets framework (panel B), edges belonging to the same circular contour pop-out from the computation similarly to a chain rule (panel C).
%This replicates the results of~\citep{Geisler01} and gives a constructive and simple implementation for this rule.
%
%Adaptive shape processing in primary visual cortex
%by: J. N. J. McManus, W. Li, C. D. Gilbert
%Proceedings of the National Academy of Sciences, Vol. 108, No. 24. (13 May 2011), pp. 9739-9746, doi:10.1073/pnas.1105855108 Key:
%Moreover, we may extend this rule to classify contours in natural images.
%Indeed, the statistics of edge co-occurences can also be computed for another class of images,
%such as images of carpentered scenes.
%Using the above rule,
%we can compute the probability for each edge belonging to a contour
%knowing one of these two classes.
%From this, we may compute an odd ratio and we show in See Figure~\ref{fig:segmentation}-B,
% the application of this rule to a natural image.
%It shows that using this rule, one can classify whether edges belong to one class or to another.
Note that contrary to classical hierarchical models,
these results are done with a simple layer of edge detection filters which communicate through local diffusion.
An important novelty to note in this extension of the SparseLets framework is that there is no recursive propagation, as the greedy algorithm is applied in a sequential manner.
These types of interaction have been found in area V1.
Indeed, the processing may be modulated by simple contextual rules
such as favoring co-linear versus co-circular edges~\citep{McManus11}.
Such type of modulation opens a wide range of potential applications to computer vision such as robust segmentation and algorithms for the autonomous classification of images~\citep{PerrinetBednar15}.
More generally, it shows that simple feed-forward algorithms such as the one we propose may be sufficient to a	account for the sparse representation of images in lower-level visual areas.
%adpatation :
% - Learning to Link Visual Contours [Quick Edit] [CiTO]
%Neuron, Vol. 57, No. 3. (7 February 2008), pp. 442-451, doi:10.1016/j.neuron.2007.12.011
%y Wu Li, Valentin Pi�ch, Charles D. Gilbert
% - learning to extract pathes in the field type of display : Natural scene statistics and the structure of orientation maps in the visual cortex Hunt09

\section{Conclusion}
%------------------%

%
%  1.  More memories can be stored
%  2.  Makes use of the statistical structure of natural signals
%  3.  Represents data in a convenient way for further processing
%  4.  Save metabolic energy, by decreasing neuronal firing rates
%

In this chapter, we have shown sparse models at increasing structural complexities mimicking the key features of the primary visual cortex in primates.
By focusing on a concrete example, the SparseLets framework, we have shown that sparse models provide 
an efficient framework for biologically-inspired computer vision algorithms.
In particular, by including contextual information, such as prior statistics on natural images, we could improve
the efficiency of the sparseness of the representation.

%Gestalt
Such an approach allows to implement a range of practical concepts (such as the good continuity of contours) in a principled way.
Indeed, we based our reasoning on inferential processes such as they are reflected in the organization of neural structures.
For instance, there is a link between co-circularity and the structure of orientation maps~\citep{Hunt09}.
This should be included in further perspectives of these sparse models.

As we saw, the (visual) brain is not a computer.
Instead of using a sequential stream of semantic symbols, it uses statistical regularities to derive predictive rules.
These computations are not written explicitly, as it suffices that they emerge from the collective behavior in populations of neurons.
As such, these rules are massively parallel, asynchronous and error prone.
Luckily, such neuromorphic computing architectures begin to emerge --- yet, we lack a better understanding of how we may implement computer vision algorithms on such hardware.

%
%Another interesting point this paper makes, is that some experimental results show neurons with much higher firing rates than would be predicted by metabolic estimates. Because these experiments often involve searching for firing neurons with an electrode, we may be systematically biasing our studies towards a minority of neurons that fires ?less sparsely? than the general populations. Solutions to this bias include chronically implanting electrodes where the positioning is set anatomically or using antidromic stimulation to identify neurons as opposed to stimulus elicited firing.
%
%Hunt09
%principled approach?
%generative models / motion Cadieu11 / Lie groups
%free energy and the measure of measure
As a conclusion, this drives the need for more biologically-driven computer vision algorithm and of a better understanding of V1.
However, such knowledge is largely incomplete~\citep{Olshausen05} and we need to develop a better understanding of results from electro-physiology.
A promising approach in that sense is to include model-driven stimulation of physiological studies~\citep{Leon12,Simoncini12} as they systematically test neural computations for a given visual task.

\subsection*{Acknowledgments} %
\Acknowledgments %
\bibliography{biblio_sparse}%

\backmatter

\end{document}